\documentclass{article}
\pdfoutput=1

\usepackage{microtype}
\usepackage{graphicx}
\usepackage{booktabs} 

\usepackage{hyperref}

\usepackage[accepted]{icml2019}
\usepackage{hos}
\usepackage{mathtools}

\def\floor#1{\lfloor #1 \rfloor}
\usepackage{tabularx} 
\usepackage[capitalize]{cleveref}
\usepackage{makecell}

\icmltitlerunning{EMI: Exploration with Mutual Information}

\begin{document}

\twocolumn[
\icmltitle{EMI: Exploration with Mutual Information}



\icmlsetsymbol{equal}{*}

\begin{icmlauthorlist}
\icmlauthor{Hyoungseok Kim}{equal,snu,nprc}
\icmlauthor{Jaekyeom Kim}{equal,snu,nprc}
\icmlauthor{Yeonwoo Jeong}{snu,nprc}
\icmlauthor{Sergey Levine}{bc}
\icmlauthor{Hyun Oh Song}{snu,nprc}
\end{icmlauthorlist}

\icmlaffiliation{snu}{Seoul National University, Department of Computer Science and Engineering}
\icmlaffiliation{bc}{UC Berkeley, Department of Electrical Engineering and Computer Sciences}
\icmlaffiliation{nprc}{Neural Processing Research Center}

\icmlcorrespondingauthor{Hyun Oh Song}{hyunoh@snu.ac.kr}

\icmlkeywords{Reinforcement learning, Representation learning}

\vskip 0.3in
]



\printAffiliationsAndNotice{\icmlEqualContribution} 

\begin{abstract}
Reinforcement learning algorithms struggle when the reward signal is very sparse. In these cases, naive random exploration methods essentially rely on a random walk to stumble onto a rewarding state. Recent works utilize intrinsic motivation to guide the exploration via generative models, predictive forward models, or discriminative modeling of novelty. We propose EMI, which is an exploration method that constructs embedding representation of states and actions that does not rely on generative decoding of the full observation but extracts predictive signals that can be used to guide exploration based on forward prediction in the representation space. Our experiments show competitive results on challenging locomotion tasks with continuous control and on image-based exploration tasks with discrete actions on Atari. The source code is available at \href{https://github.com/snu-mllab/EMI}{https://github.com/snu-mllab/EMI}.
\end{abstract}

\vspace{-1em}
\section{Introduction}
The central task in reinforcement learning is to learn policies that would maximize the total reward received from interacting with the unknown environment. Although recent methods have been demonstrated to solve a range of complex tasks \citep{dqn_nature, trpo, ppo}, the success of these methods hinges on whether the agent constantly receives the intermediate reward feedback or not. In case of challenging environments with sparse reward signals, these methods struggle to obtain meaningful policies unless the agent luckily stumbles into the rewarding or predefined goal states.

To this end, prior works on exploration generally utilize some kind of intrinsic motivation mechanism to provide a measure of novelty. These measures can be based on density estimation via generative models~\citep{bellemare16,ex2,action_conditional}, predictive forward models~\citep{stadie,vime}, or discriminative methods that aim to approximate novelty~\citep{icm}. Methods based on predictive forward models and generative models must model the distribution over state observations, which can make them difficult to scale to complex, high-dimensional observation spaces.

Our aim in this work is to devise a method for exploration that does not require a direct generation of high-dimensional state observations, while still retaining the benefits of being able to measure novelty based on the forward prediction. If exploration is performed by seeking out states that maximize surprise, the problem, in essence, is in measuring surprise, which requires a representation where functionally similar states are close together, and functionally distinct states are far apart.

\begin{figure}
\centering
\includegraphics[width=0.50\textwidth]{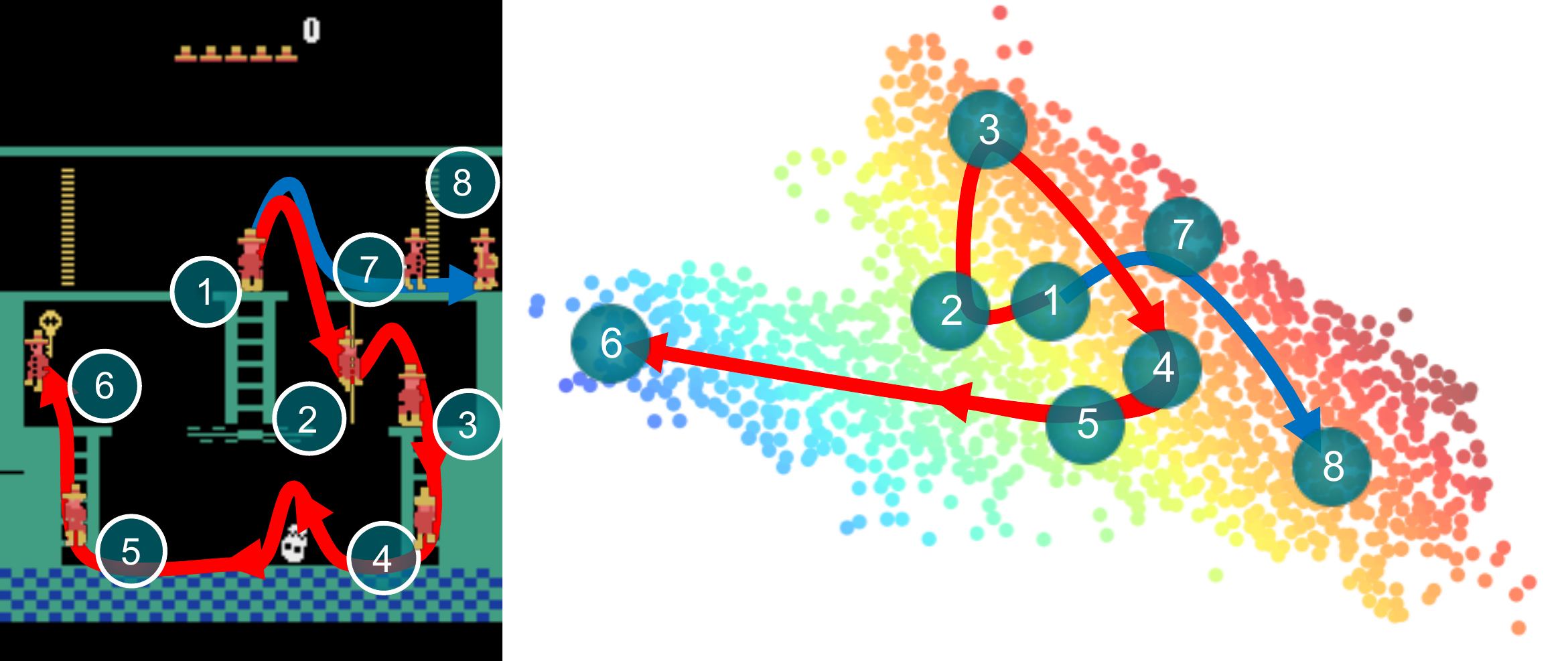}
\caption{Visualization of a sample trajectory in our learned embedding space.}
\label{fig:concept_linear_dynamics}
\vspace{-2em}
\end{figure}

In this paper, we propose to learn compact representations for both the states $(\phi)$ and actions $(\psi)$ simultaneously satisfying the following criteria: First, given the representations of state and the corresponding next state, the uncertainty of the representation of the corresponding action should be minimal. Second, given the representations of the state and the corresponding action, the uncertainty of the representation of the corresponding next state should also be minimal. Third, the action embedding representation  $(\psi)$ should seamlessly support both continuous and discrete actions. Finally, we impose a linear dynamics model in the representation space which can also explain the rare irreducible error under the dynamics model. Given the representation, we guide the exploration by measuring surprise based on forward prediction and a relative increase in diversity in the embedding representation space. \Cref{fig:concept_linear_dynamics} illustrates an example visualization of our learned state embedding representations $(\phi)$ and sample trajectories in the representation space in Montezuma's Revenge.

We present two main technical contributions that make this into a practical exploration method. First, we describe how compact state and action representations can be constructed via variational divergence estimation of mutual information without relying on generative decoding of full observations \cite{fgan}. Second, we show that imposing linear topology on the learned embedding representation space (such that the transitions are linear), thereby offloading most of the modeling burden onto the embedding function itself, provides an essential informative measure of surprise when visiting novel states.

For the experiments, we show that we can use our representations on a range of complex image-based tasks and robotic locomotion tasks with continuous actions. We report significantly improved results compared to a number of recent intrinsic motivation based exploration methods \citep{ex2,icm} on several challenging Atari tasks and robotic locomotion tasks with sparse rewards.    

\section{Related works}
Our work is related to the following strands of active research:

\textbf{Unsupervised representation learning via mutual information estimation}~ Recent literature on unsupervised representation learning generally focuses on extracting latent representations maximizing an approximate lower bound on the mutual information between the code and the data. In the context of generative adversarial networks \citep{gan}, \citet{infogan,mine} aim at maximizing the approximation of mutual information between the latent code and the raw data. \citet{mine} estimates the mutual information with neural networks via \citet{donsker1983} estimation to learn better generative models. \citet{dim} builds on the idea and trains a decoder-free encoding representation maximizing the mutual information between the input image and the representation. Furthermore, the method uses $f$-divergence \citep{fgan} estimation of Jensen-Shannon divergence rather than the KL divergence to estimate the mutual information for better numerical stability. \citet{bengio2017, thomas2017, icf} define \emph{selectivity}, which lower bounds the conditional mutual information between the embedding of the next state and the policy embedding given the embedding of the current state, in order to learn disentangled factors of variation. \citet{cpc} estimates mutual information via an autoregressive model and makes predictions on local patches in an image. \citet{nachum2018near} connects mutual information estimators to representation learning in hierarchical RL.

\textbf{Exploration with intrinsic motivation}~ Prior works on exploration mostly employ intrinsic motivation to estimate the measure of novelty or surprisal to guide the exploration. \citet{mohamed} introduced the connection between mutual information estimation and empowerment for intrinsic motivation. \citet{bellemare16,ostrovski17} utilize density estimation via CTS \citep{bellemare14} generative model and PixelCNN \cite{vandenoord16} and derive pseudo-counts as the intrinsic motivation. \citet{ex2} avoids building explicit density models by training K-exemplar models that distinguish a state from all other observed states. Some methods train predictive forward models \citep{stadie,vime,action_conditional} and estimate the prediction error as the intrinsic motivation. \citet{action_conditional} employs generative decoding of the full observation via recursive autoencoders and thus can be challenging to scale for high dimensional observations. VIME \cite{vime} approximates the environment dynamics, uses the information gain of the learned dynamics model as intrinsic rewards, and showed encouraging results on robotic locomotion problems. However, the method needs to update the dynamics model per each observation and is unlikely to be scalable for complex tasks with high dimensional states such as Atari games. 

RND \cite{rnd} trains a network to predict the output of a fixed randomly initialized target network and uses the prediction error as the intrinsic reward but the method does not report the results on continuous control tasks. ICM \cite{icm} transforms the high dimensional states to feature space and imposes cross entropy and Euclidean loss so the action and the feature of the next state are predictable. However, ICM does not utilize mutual information like VIME to directly measure the uncertainty and is limited to discrete actions. Our method (EMI) is also reminiscent of \citep{kohonen1998} in the sense that we seek to construct a decoder-free latent space from the high dimensional observation data with a topology in the latent space. In contrast to the prior works on exploration, we seek to construct the representation under linear topology and does not require decoding the full observation but seek to encode the essential predictive signal that can be used for guiding the exploration.

\section{Preliminaries}
We consider a Markov decision process defined by the tuple $(\mathcal{S}, \mathcal{A}, P, r, \gamma)$, where $\mathcal{S}$ is the set of states, $\mathcal{A}$ is the set of actions, $P:  \mathcal{S} \times \mathcal{A} \times \mathcal{S} \rightarrow \reals_+$ is the environment transition distribution, $r: \mathcal{S} \rightarrow \reals$ is the reward function, and $\gamma \in (0,1)$ is the discount factor. Let $\pi$ denote a stochastic policy over actions given states. Denote $\mathbb{P}_0: \mathcal{S} \rightarrow \reals_+$ as the distribution of initial state $s_0$. The discounted sum of expected rewards under the policy $\pi$ is defined by 
\begin{align*}
\eta(\pi) = \mathbb{E}_\tau \left[ \sum_{t=0} \gamma^t r(s_t)\right],
\end{align*}
where $\tau = (s_0, a_0, \ldots, a_{T-1}, s_T)$ denotes the trajectory, $s_0 \sim \mathbb{P}_0(s_0), a_t \sim \pi(a_t \mid s_t),$ and $s_{t+1} \sim P(s_{t+1} \mid s_t, a_t)$. The objective in policy based reinforcement learning is to search over the space of parameterized policies (\ie~ neural network) $\pi_\theta(a \mid s)$ in order to maximize $\eta(\pi_\theta)$.

Also, denote $\mathbb{P}_{SAS'}^\pi$ as the joint probability distribution of singleton  experience tuples $(s, a, s')$ starting from $s_0 \sim \mathbb{P}_0(s_0)$ and following the policy $\pi$. Furthermore, define $\mathbb{P}_A^\pi = \int_{\mathcal{S} \times \mathcal{S}'} d\mathbb{P}_{SAS'}^\pi$ as the marginal distribution of actions, $\mathbb{P}_{SS'}^\pi = \int_{\mathcal{A}} d\mathbb{P}_{SAS'}^\pi$ as the marginal distribution of states and the corresponding next states, $\mathbb{P}_{S'}^\pi = \int_{\mathcal{S} \times \mathcal{A}} d\mathbb{P}_{SAS'}^\pi$ as the marginal distribution of the next states, and $\mathbb{P}_{SA}^\pi = \int_{\mathcal{S}'} d\mathbb{P}_{SAS'}^\pi$ as the marginal distribution of states and the actions following the policy $\pi$.

\section{Methods}
Our goal is to construct the embedding representation of the observation and action (discrete or continuous) for complex dynamical systems that does not rely on generative decoding of the full observation, but still provides a useful predictive signal that can be used for exploration. This requires a representation where functionally similar states are close together, and functionally distinct states are far apart. We approach this objective from the standpoint of maximizing mutual information under several criteria.

\subsection{Mutual information maximizing state and action embedding representations} \label{sec:mi}

In this subsection, we introduce the desiderata for our objective and discuss the variational divergence lower bound for efficient computation of the objective. We denote the embedding function of states $\phi_\alpha : \mathcal{S} \rightarrow \reals^d$ and actions $\psi_\beta: \mathcal{A} \rightarrow \reals^d$ with parameters $\alpha$ and $\beta$ (\ie~ neural networks) respectively. We seek to learn the embedding function of states ($\phi_\alpha$) and actions ($\psi_\beta$) satisfying the following two criteria: 

\vspace{-0.25cm}
\begin{enumerate}
\item Given the embedding representation of states and the actions $[\phi_\alpha(s); \psi_\beta(a)]$, the uncertainty of the embedding representation of the corresponding next states $\phi_\alpha(s')$ should be minimal and vice versa.

\item Given the embedding representation of states and the corresponding next states $[\phi_\alpha(s); \phi_\alpha(s')]$, the uncertainty of the embedding representation of the corresponding actions $\psi_\beta(a)$ should also be minimal and vice versa.

\end{enumerate}
\vspace{-0.25cm}

Intuitively, the first criterion translates to maximizing the mutual information between $[\phi_\alpha(s); \psi_\beta(a)], $ and $\phi_\alpha(s')$ which we define as $\mathcal{I}_S(\alpha, \beta)$ in \Cref{eqn:infogain_s}. And the second criterion translates to maximizing the mutual information between $[\phi_\alpha(s); \phi_\alpha(s')]$ and $\psi_\beta(a)$ defined as $\mathcal{I}_A(\alpha, \beta)$ in \Cref{eqn:infogain_a}. 
\vspace{-0.2em}
\begin{align}
\maximize_{\alpha, \beta}~ \mathcal{I}_S(\alpha, \beta) &:= \mathcal{I}([\phi_\alpha(s); \psi_\beta(a)]; \phi_\alpha(s')) \nonumber \\
&= \mathcal{D}_{\text{KL}}\left(\mathbb{P}_{SAS'}^\pi \parallel \mathbb{P}_{SA}^\pi \otimes \mathbb{P}_{S'}^\pi\right)
\label{eqn:infogain_s}
\end{align}
\vspace{-1em}
\begin{align}
\maximize_{\alpha, \beta}~ \mathcal{I}_A(\alpha, \beta) &:= \mathcal{I}([\phi_\alpha(s); \phi_\alpha(s')]; \psi_\beta(a)) \nonumber \\
&= \mathcal{D}_{\text{KL}}\left(\mathbb{P}_{SAS'}^\pi \parallel \mathbb{P}_{SS'}^\pi \otimes \mathbb{P}_A^\pi\right)
\label{eqn:infogain_a}
\end{align}
\vspace{-0.5cm}

\begin{figure*}[ht]
    \centering
    \includegraphics[width=1\textwidth]{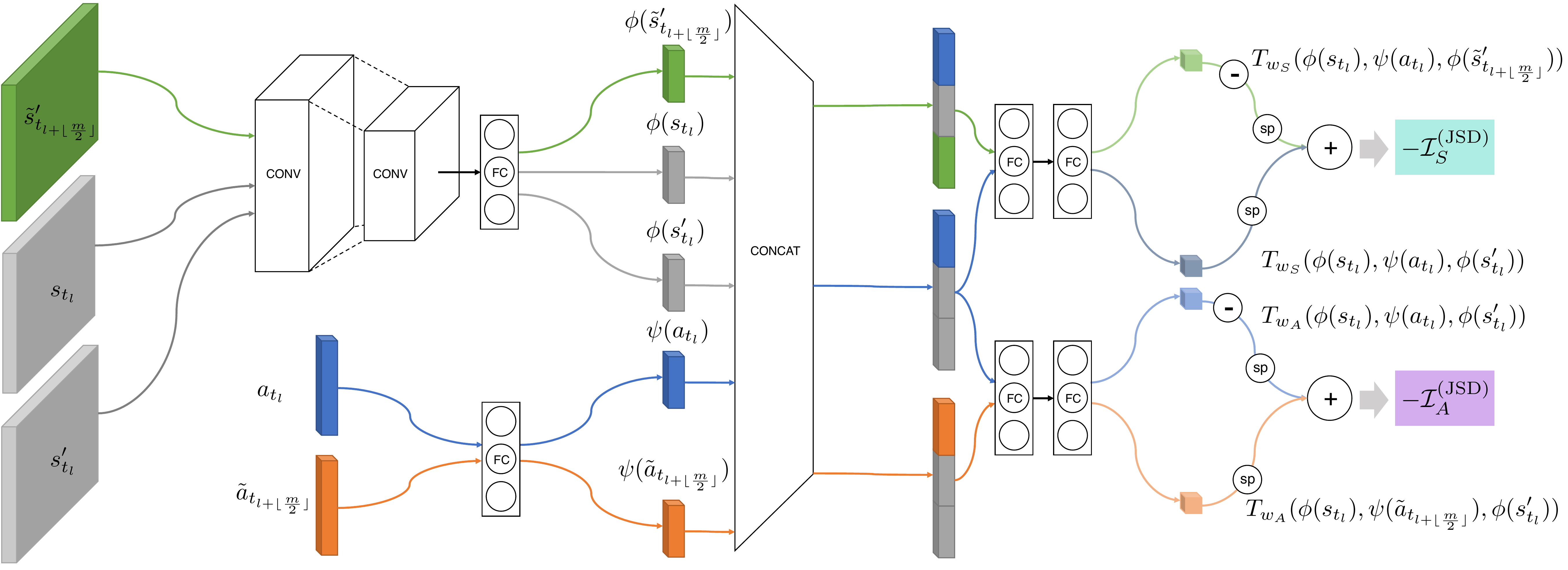}
    \caption{Computational architecture for estimating $\mathcal{I}^{\text{(JSD)}}_S$ and $\mathcal{I}^{\text{(JSD)}}_A$ for image-based observations.}
    \label{fig:architecture}
    \vspace{-0.5em}
\end{figure*}

Mutual information is not bounded from above and maximizing mutual information is notoriously difficult to compute in high dimensional settings. Motivated by \cite{dim, mine}, we compute the variational divergence lower bound of mutual information \cite{fgan}. Concretely, variational divergence (f-divergence) representation is a tight estimator for the mutual information of two random variables $X$ and $Z$, derived as in \Cref{eqn:donsker_kl}.

\vspace{-1.2em}
\begin{align}
\label{eqn:donsker_kl}
&\mathcal{I}(X; Z) = \mathcal{D}_\text{KL}(\mathbb{P}_{XZ} \parallel \mathbb{P}_{X} \otimes \mathbb{P}_Z)  \\
&\qquad \geq \sup_{\omega \in \Omega} \mathbb{E}_{\mathbb{P}_{XZ}} T_\omega(x, z) - \log \mathbb{E}_{\mathbb{P}_{X} \otimes \mathbb{P}_Z} \exp(T_\omega(x, z)), \nonumber
\end{align}

\vspace{-1em}

where $T_\omega : \mathcal{X} \times \mathcal{Z} \rightarrow \reals$ is a differentiable transform with parameter $\omega$. Furthermore, for better numerical stability, we utilize a different measure between the joint and marginals than the KL-divergence. In particular, we employ Jensen-Shannon divergence (JSD) \citep{dim} which is bounded both from below and above by $0$ and $\log(4)$ \footnote{In \cite{fgan}, the authors derive the lower bound of $D_{JSD} = D_{KL}(P||M)+D_{KL}(Q||M)$, instead of $D_{JSD} =\frac{1}{2}(D_{KL}(P||M)+D_{KL}(Q||M))$, where $M= \frac{1}{2} (P + Q).$}. 

\vspace{0.5em}

\begin{theorem}
The lower bound of mutual information using Jensen-Shannon divergence is
\begin{align*}
    \mathcal{I}^{(\text{JSD})}(X; Z) &\geq \sup_{\omega \in \Omega} 
    \mathbb{E}_{\mathbb{P}_{XZ}} \left[-\text{sp} \left(-T_{\omega}(x,z)\right)\right] \\
    & - \mathbb{E}_{\mathbb{P}_{X} \otimes \mathbb{P}_Z} \left[ \text{sp} \left(T_{\omega}(x,z)\right)\right]+\log(4)
\end{align*}
\label{thm:JSD}
\end{theorem}

\vspace{-3em}

\begin{proof}
\begin{align*}
    &\mathcal{I}^{(\text{JSD})}(X; Z) = \mathcal{D}_\text{JSD}(\mathbb{P}_{XZ} \parallel \mathbb{P}_{X} \otimes \mathbb{P}_Z)\nonumber\\
    &\geq \sup_{\omega \in \Omega} \mathbb{E}_{\mathbb{P}_{XZ}} \left[S_\omega(x,z)\right] - \mathbb{E}_{\mathbb{P}_{X} \otimes \mathbb{P}_Z} \left[\text{JSD}^{*}\left(S_\omega(x,z)\right)\right] \nonumber\\
    &= \sup_{\omega \in \Omega} \mathbb{E}_{\mathbb{P}_{XZ}} \left[-\text{sp} \left(-T_{\omega}(x,z)\right)\right] \nonumber \\ 
    & \qquad \quad - \mathbb{E}_{\mathbb{P}_{X} \otimes \mathbb{P}_Z} \left[ \text{sp} \left(T_{\omega}(x,z)\right)\right] +\log(4),
\end{align*}
where the inequality in the second line holds from the definition of $f$-divergence \citep{fgan}. In the third line, we substituted $S_\omega(x,z) = \log(2) - \log(1+\exp(-T_\omega(x,z)))$ and \textit{Fenchel conjugate} of Jensen-Shannon divergence, $\text{JSD}^{*}(t) = -\log(2-\exp(t))$.
\end{proof}

From \Cref{thm:JSD}, we have,

\vspace{-2em}
\fontsize{8pt}{1em}\selectfont
\begin{align} 
\label{eqn:infogain_s_jsd}
&\maximize_{\alpha, \beta}~ \mathcal{I}^{\text{(JSD)}}_S(\alpha, \beta) \nonumber \\
&\geq \maximize_{\alpha, \beta}  \sup_{\omega_S \in \Omega_S}\mathbb{E}_{\mathbb{P}_{SAS'}^\pi} \left[-\text{sp} \left(-T_{\omega_S}(\phi_\alpha(s), \psi_\beta(a), \phi_\alpha(s'))\right)\right] \nonumber \\
& -\mathbb{E}_{\mathbb{P}_{SA}^\pi \otimes \mathbb{P}_{S'}^\pi} \left[ \text{sp} \left(T_{\omega_S}(\phi_\alpha(s), \psi_\beta(a), \phi_\alpha(\tilde{s'}))\right)\right] + \log{4},
\end{align}

\vspace{-2em}

\begin{align}
\label{eqn:infogain_a_jsd}
&\maximize_{\alpha, \beta}~ \mathcal{I}^{\text{(JSD)}}_A(\alpha, \beta) \nonumber \\
&\geq \maximize_{\alpha, \beta}  \sup_{\omega_A \in \Omega_A} \mathbb{E}_{\mathbb{P}_{SAS'}^\pi} \left[-\text{sp} \left(-T_{\omega_A}(\phi_\alpha(s), \psi_\beta(a), \phi_\alpha(s'))\right)\right] \nonumber \\
& -\mathbb{E}_{\mathbb{P}_{SS'}^\pi \otimes \mathbb{P}_{A}^\pi} \left[ \text{sp} \left(T_{\omega_A}(\phi_\alpha(s), \psi_\beta(\tilde{a}), \phi_\alpha(s'))\right)\right] +\log{4},
\end{align}
\normalsize
\vspace{-0.5cm}

where $\text{sp}(z)=\log(1 + \exp(z))$. The expectations in \Cref{eqn:infogain_s_jsd} and \Cref{eqn:infogain_a_jsd} are approximated using the empirical samples trajectories $\tau$. Note, the samples $\tilde{s'} \sim \mathbb{P}_{S'}^\pi$ and $\tilde{a} \sim \mathbb{P}_A^\pi$ from the marginals are obtained by dropping $(s,a)$ and $(s,s')$ in samples $(s, a, \tilde{s'})$ and $(s, \tilde{a}, s')$ from $\mathbb{P}_{SAS'}^\pi$. \Cref{fig:architecture} illustrates the computational architecture for estimating the lower bounds on $\mathcal{I}_S$ and $\mathcal{I}_A$. 

\subsection{Embedding the linear dynamics model with the error model}

Since the embedding representation space is learned, it is natural to impose a topology on it \citep{kohonen1983}. In EMI, we impose a simple and convenient topology where transitions are linear since this spares us from having to also represent a complex dynamical model. This allows us to offload most of the modeling burden onto the embedding function itself, which in turn provides us with a useful and informative measure of surprise when visiting novel states. Once the embedding representations are learned, this linear dynamics model allows us to measure \emph{surprise} in terms of the residual error under the model or measure \emph{diversity} in terms of the similarity in the embedding space. \Cref{sec:intrinsic} discusses the intrinsic reward computation procedure in more detail.

Concretely, we seek to learn the representation of states $\phi(s)$ and the actions $\psi(a)$ such that the representation of the corresponding next state $\phi(s')$ follow linear dynamics \ie~ $\phi(s') = \phi(s) + \psi(a)$. Intuitively, we would like the nonlinear aspects of the dynamics to be offloaded to the neural networks $\phi(\cdot), \psi(\cdot)$ so that in the $\reals^d$ embedding space, the dynamics become linear. Regardless of the expressivity of the neural networks, however, there always exists irreducible error under the linear dynamic model. For example, the state transition which leads the agent from one room to another in Atari environments (\ie~ Venture, Montezuma's revenge, \etc) or the transition leading the agent in the same position under certain actions (\ie~ Agent bumping into a wall when navigating a maze) would be extremely challenging to explain under the linear dynamics model. 

To this end, we introduce the error model $S_\gamma : \mathcal{S} \times \mathcal{A} \rightarrow \reals^d$, which is another neural network taking the state and action as input, estimating the irreducible error under the linear model. Motivated by the work of \citet{robustpca}, we seek to minimize Frobenius norm of the error term so that the error term contributes on sparingly unexplainable occasions. \Cref{eqn:sparse} shows the embedding learning problem under linear dynamics with modeled errors.

\vspace{-2em}

\begin{align}
&\minimize_{\alpha, \beta, \gamma} \underbrace{\| S_\gamma \|_{2,0}}_{\text{error minimization}} \nonumber\\
&\text{\ subject to }~~ \underbrace{\Phi_{\alpha}' = \Phi_\alpha + \Psi_\beta + S_\gamma}_{\text{embedding linear dynamics}}, 
\label{eqn:sparse}
\end{align}


where we used the matrix notation for compactness. $\Phi_\alpha, \Psi_\beta, S_\gamma$ denotes the matrices of respective embedding representations stacked columns wise. Relaxing the matrix $\|\cdot\|_{2,0}$ norm with Frobenius norm, \Cref{eqn:master_eqn} shows our final learning objective.

\vspace{-1.5em}

\begin{align} \label{eqn:master_eqn}
\minimize_{\alpha, \beta, \gamma} ~ &\|\Phi'_\alpha - \left( \Phi_\alpha + \Psi_\beta  + S_\gamma \right)\|_F^2 \nonumber\\
&+ \lambda_{\text{error}} \| S_\gamma \|_F^2
+ \lambda_\text{info} \mathcal{L}_\text{info},
\end{align}
where $\mathcal{L}_\text{info}$ denotes the following mutual information term.
\begin{align}
\mathcal{L}_\text{info} &= \inf_{\omega_S \in \Omega_S} \mathbb{E}_{\mathbb{P}_{SAS'}^\pi} \text{sp} \left(-T_{\omega_S}(\phi_\alpha(s), \psi_\beta(a), \phi_\alpha(s'))\right) \nonumber \\
&\qquad \quad + \mathbb{E}_{\mathbb{P}_{SA}^\pi \otimes \mathbb{P}_{S'}^\pi} ~\text{sp} \left(T_{\omega_S}(\phi_\alpha(s), \psi_\beta(a), \phi_\alpha(\tilde{s'}))\right) \nonumber \\
&+ \inf_{\omega_A \in \Omega_A} \mathbb{E}_{\mathbb{P}_{SAS'}^\pi} \text{sp} \left(-T_{\omega_A}(\phi_\alpha(s), \psi_\beta(a), \phi_\alpha(s'))\right) \nonumber \\
&\qquad \quad + \mathbb{E}_{\mathbb{P}_{SS'}^\pi \otimes \mathbb{P}_{A}^\pi} ~\text{sp} \left(T_{\omega_A}(\phi_\alpha(s), \psi_\beta(\tilde{a}), \phi_\alpha(s'))\right) \nonumber
\end{align}
\normalsize
$\lambda_\text{error}, \lambda_\text{info}$ are hyperparameters which control the relative contributions of the linear dynamics error and the mutual information term. In practice, for image-based experiments, we found the optimization process to be more stable when we further regularize the distribution of action embedding representation to follow a predefined prior distribution. Concretely, we regularize the action embedding distribution to follow a standard normal distribution via $  \mathcal{D}_\text{KL}(\mathbb{P}^\pi_{\psi} \parallel \mathcal{N}(0,I))$ similar to VAEs \cite{vae}. Intuitively, this has the effect of grounding the distribution of action embedding representation (and consequently the state embedding representation) across different iterations of the learning process. 

Note, regularizing the distribution of \emph{state} instead of \emph{action} embeddings renders the optimization process much more unstable. This is because the distribution of states are much more likely to be skewed than the distribution of actions, especially during the initial stage of optimization, so the Gaussian approximation becomes much less accurate in contrast to the distribution of actions. In \Cref{sec:regularization_target}, we compare the state and action embeddings as regularization targets in terms of the quality of the learned embedding functions.

\subsection{Intrinsic reward augmentation} \label{sec:intrinsic}
We consider a formulation based on the prediction error under the linear dynamics model as shown in \Cref{eqn:ir_err}. This formulation incorporates the error term and makes sure we differentiate the irreducible error that does not contribute as the novelty.

\vspace{-2em}

\begin{align}
\hspace{-0.5em}  r_e(s_t, a_t, s_t') = \|\phi(s_t) + \psi(a_t) + S(s_t, a_t) - \phi(s_t')\|^2
\label{eqn:ir_err}
\end{align}

\vspace{-1em}

\Cref{alg:procedure} shows the complete procedure in detail. The choice of different intrinsic reward formulation and the computation of $\mathcal{L}_\text{info}$ are fully described in supplementary Section 2 and 3.

\begin{algorithm}[H]
   \caption{Exploration with mutual information state and action embeddings (EMI)}
   \label{alg:procedure}
\begin{algorithmic}
    \REQUIRE $\alpha, \beta, \gamma, \omega_A, \omega_S$
    \FOR{ $i=1,\ldots,$ MAXITER}
        \STATE Collect samples $\{(s_t, a_t, s_t') \}_{t=1}^n$ with policy $\pi_\theta$
        \STATE Compute prediction error intrinsic rewards $\{r_e(s_t, a_t, s_t')\}_{t=1}^n$ following \Cref{eqn:ir_err}
        \FOR{ $j=1,\ldots,$ OPTITER}
            \FOR{ $k=1,\ldots, \lfloor \frac{n}{m} \rfloor$}
                \STATE Sample a minibatch $\{(s_{t_l}, a_{t_l}, s_{t_l}') \}_{l=1}^m$
                \STATE Update $\alpha, \beta, \gamma, \omega_A, \omega_S$ using the Adam update rule to minimize \Cref{eqn:master_eqn}
            \ENDFOR
        \ENDFOR
        \STATE Augment the intrinsic rewards with environment reward $r_\text{env}$ as $r = r_\text{env} + \eta r_e$ and update the policy network $\pi_\theta$ using any RL method
    \ENDFOR
\end{algorithmic}
\end{algorithm}
\vspace{-1em}

\begin{figure*}[ht]
    \centering

    \begin{subfigure}[t]{0.85\textwidth}
        \centering
        \includegraphics[width=0.85\textwidth]{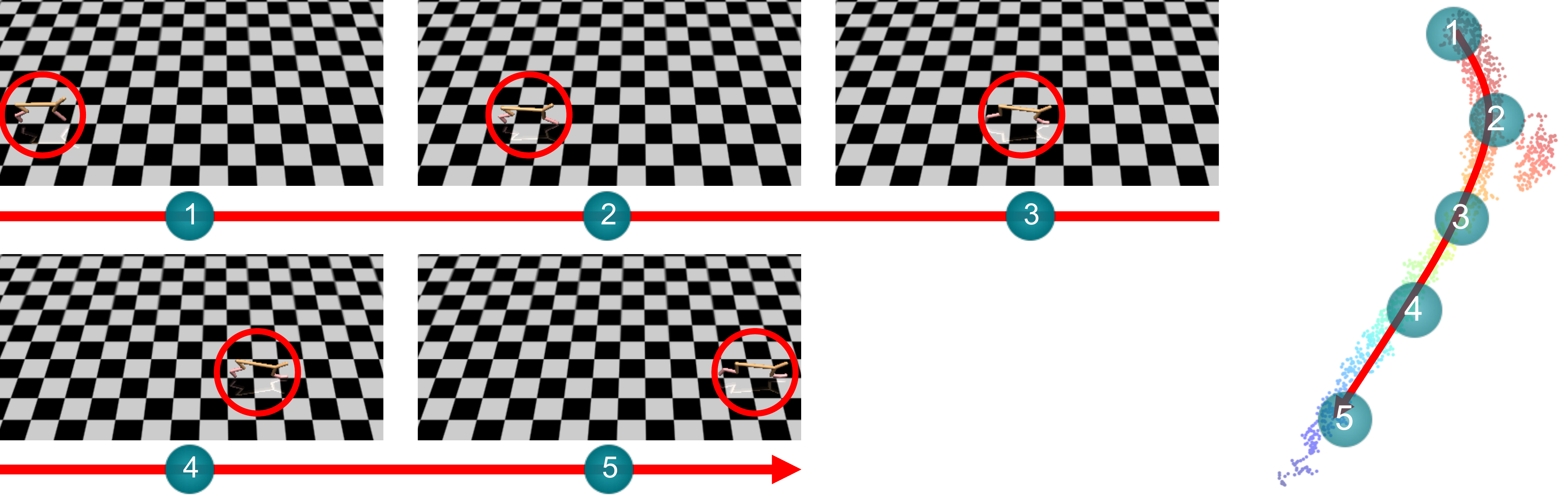}
        \caption{Example paths in and our state embeddings for SparseHalfCheetah}
        \label{fig:paths_emb_sparsehalfcheetah}
    \end{subfigure}\\
    
    \begin{subfigure}[t]{0.85\textwidth}
        \centering
        \includegraphics[width=0.85\textwidth]{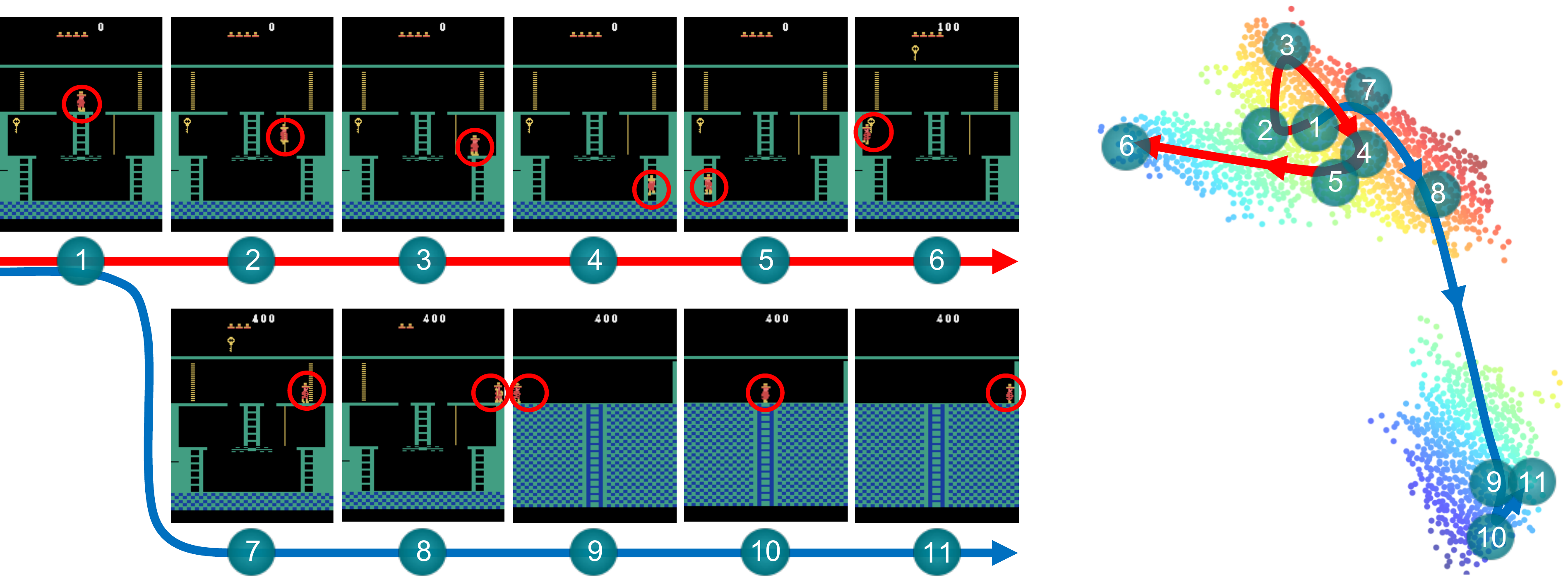}
        \caption{Example paths in and our state embeddings for Montezuma's Revenge}
        \label{fig:paths_emb_montezuma}
    \end{subfigure}\\
    
    \begin{subfigure}[t]{0.85\textwidth}
        \centering
        \includegraphics[width=0.85\textwidth]{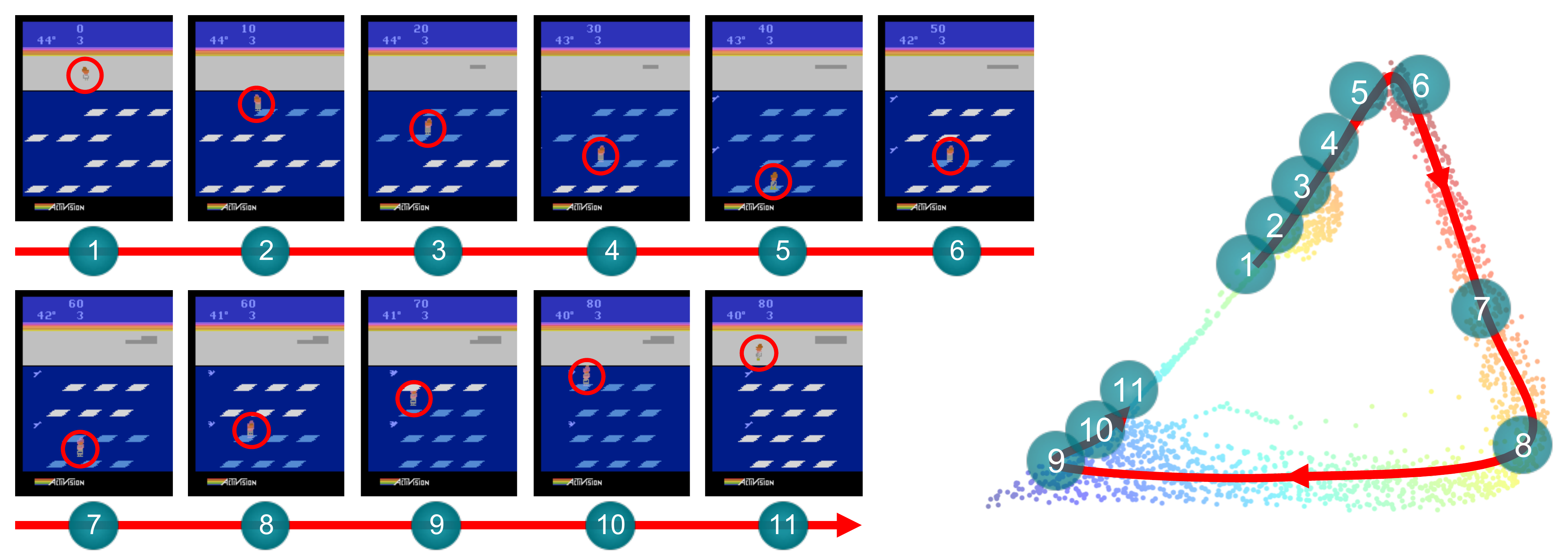}
        \caption{Example paths in and our state embeddings for Frostbite}
        \label{fig:paths_emb_frostbite}
    \end{subfigure}
    
    \caption{Example sample paths in our learned embedding representations. Note the embedding dimensionality $d$ is $2$, and thus we did not use any dimensionality reduction techniques.}
    \label{fig:obs_emb_paths}
    \vspace{-1em}
\end{figure*}

\vspace{-1em}

\section{Experiments}
We compare the experimental performance of EMI to recent prior works on both low-dimensional locomotion tasks with continuous control from rllab benchmark \citep{rllab} and the complex vision-based tasks with discrete control from the Arcade Learning Environment \citep{ale}. For the locomotion tasks, we chose SwimmerGather and SparseHalfCheetah environments for direct comparison against the prior work of \cite{ex2}. SwimmerGather is a hierarchical task where a two-link robot needs to reach green pellets, which give positive rewards, instead of red pellets, which give negative rewards. SparseHalfCheetah is a challenging locomotion task where a cheetah-like robot does not receive any rewards until it moves 5 units in one direction.

For vision-based tasks, we selected Freeway, Frostbite, Venture, Montezuma's Revenge, Gravitar, and Solaris for comparison with recent prior works \citep{icm,ex2,rnd}. These six Atari environments feature very sparse reward feedback and often contain many moving distractor objects which can be challenging for the methods that rely on explicit decoding of the full observations \citep{action_conditional}. \Cref{table:compare} shows the overall performance of EMI compared to the baseline methods in all tasks.

\subsection{Implementation Details}
We compare all exploration methods using the same RL procedure, in order to provide a fair comparison. Specifically, we use TRPO~\cite{trpo}, a policy gradient method that can be applied to both continuous and discrete action spaces. Although the absolute performance on each task depends strongly on the choice of RL algorithm, comparing the different methods with the same RL procedure allows us to control for this source of variability. Also, we observed TRPO is less sensitive to changes in hyperparameters than A3C (see \citet{a3c}) making the comparisons easier.

In the locomotion experiments, we use a 2-layer fully connected neural network as the policy network. In the Atari experiments, we use a 2-layer convolutional neural network followed by a single layer fully connected neural network. We convert the 84 x 84 input RGB frames to grayscale images and resize them to 52 x 52 images following the practice in \citet{hash_exploration}. The embedding dimensionality is set to $d=2$ and intrinsic reward coefficient is set to $\eta=0.001$ in all of the environments. We use Adam \citep{adam} optimizer to train embedding networks. Please refer to supplementary Section 1 for more details.

\begin{figure*}[ht!]
    \centering
    \begin{subfigure}{0.32\textwidth}
        \includegraphics[width=\textwidth]{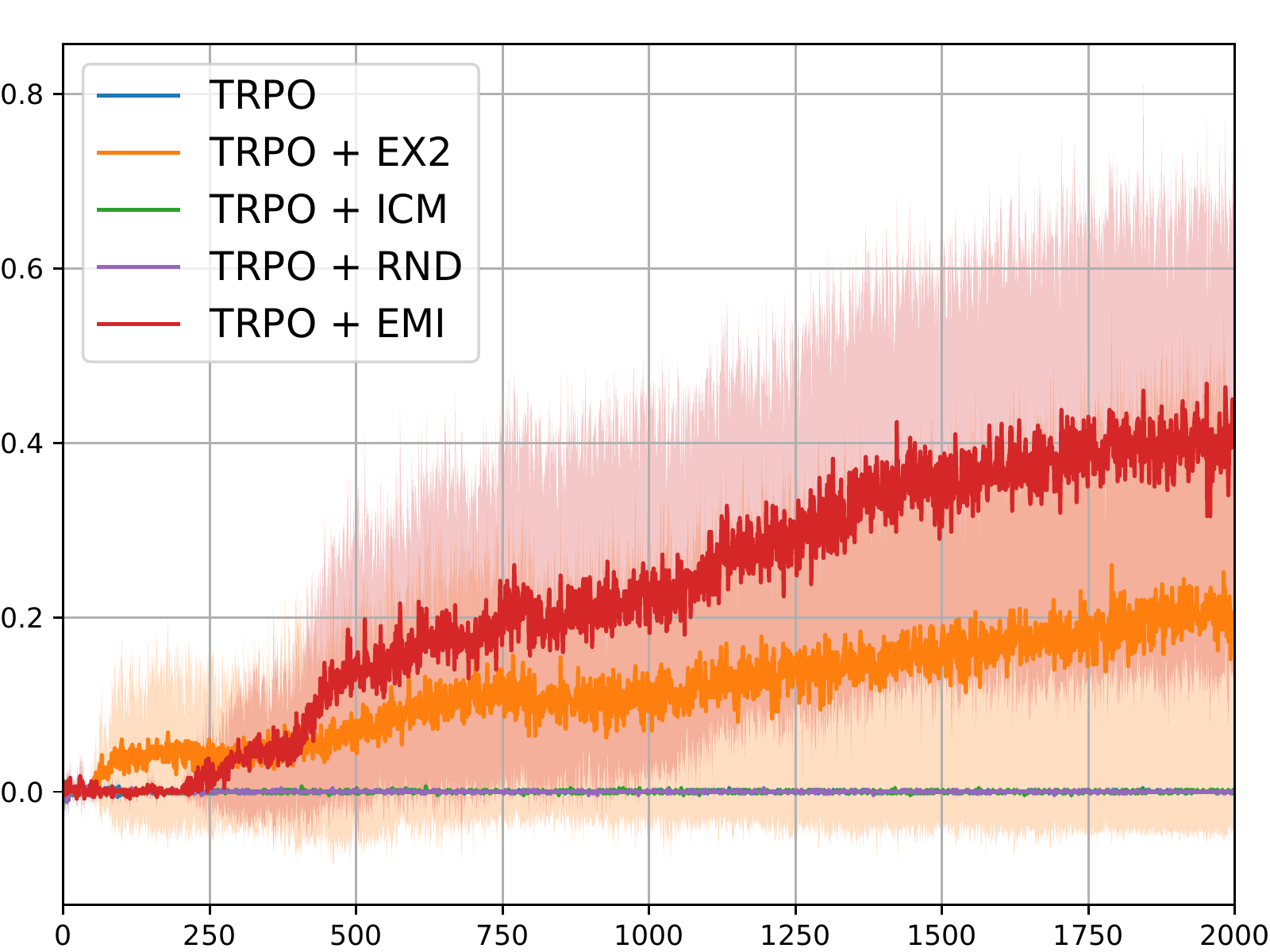}
        \caption{SwimmerGather}
        \label{fig:SwimmerGather_returns}
    \end{subfigure}
    \begin{subfigure}{0.32\textwidth}
        \includegraphics[width=\textwidth]{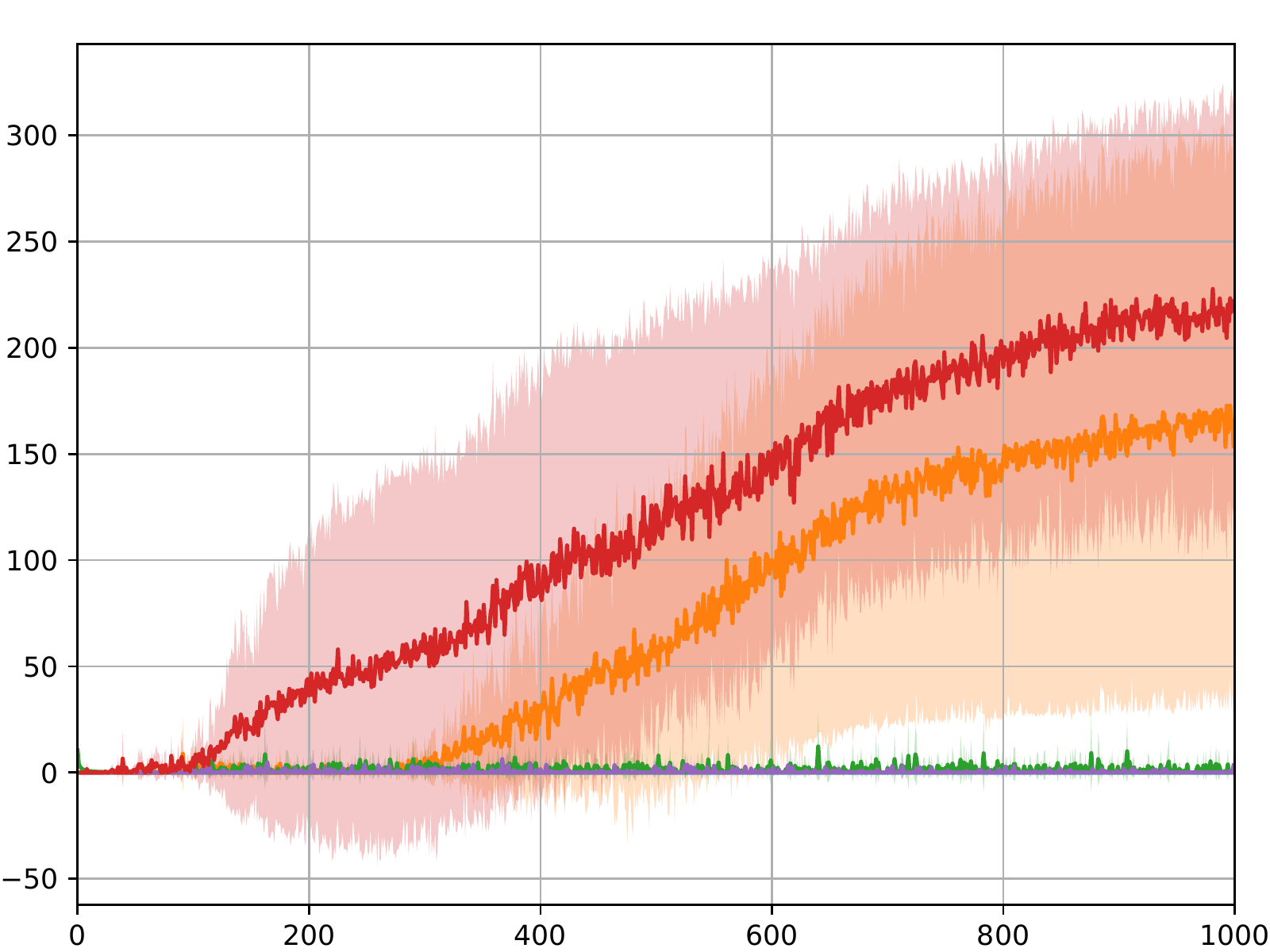}
        \caption{SparseHalfCheetah}
        \label{fig:SparseHalfCheetah_returns}
    \end{subfigure}
     \begin{subfigure}{0.32\textwidth}
        \includegraphics[width=\textwidth]{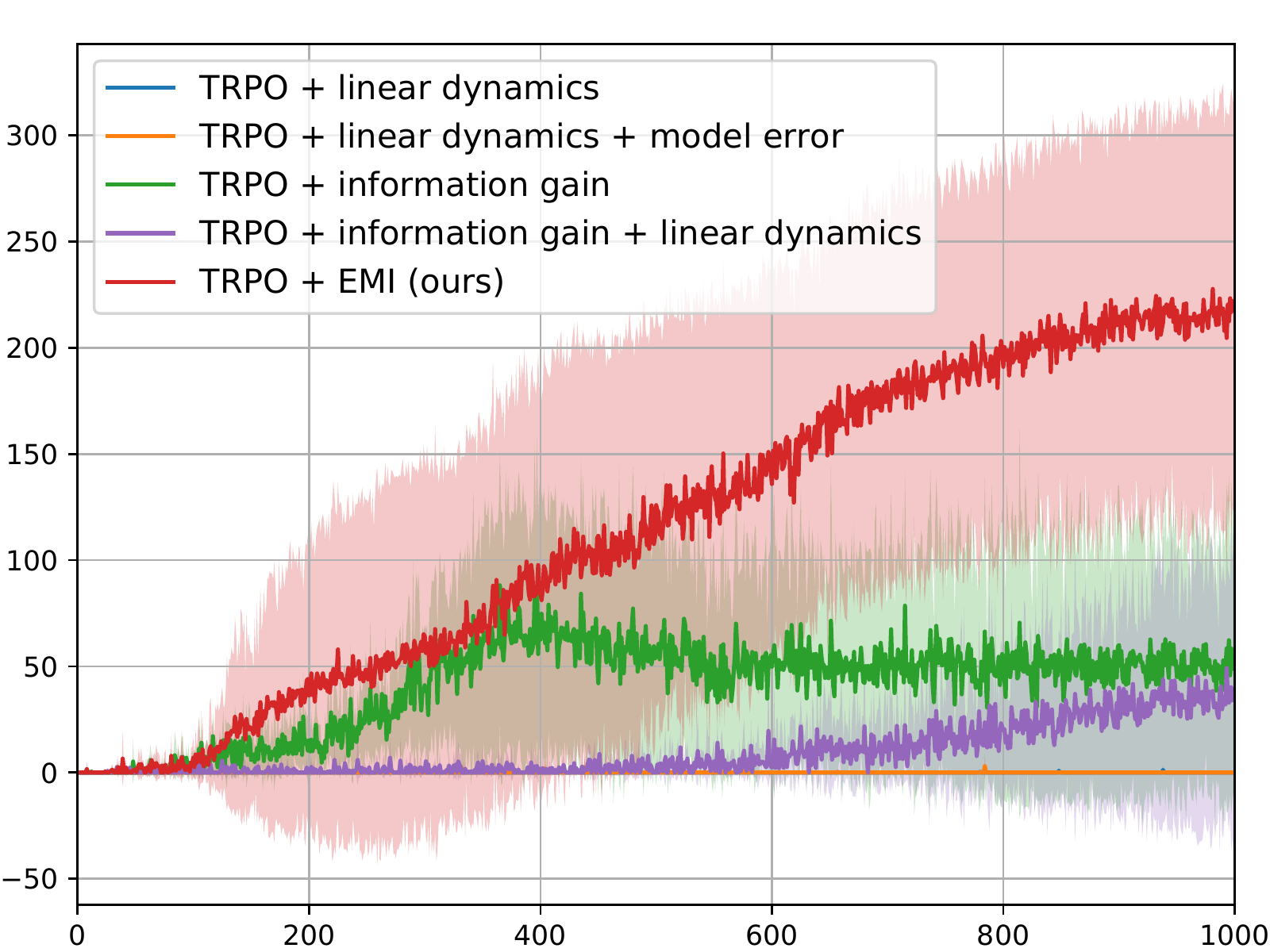}
    	\caption{Ablation study}
	\label{fig:mujoco_ablation}
    \end{subfigure}
    \caption{(a), (b): Performance of EMI on locomotion tasks with sparse rewards compared to the baseline methods. The solid lines show the mean reward (y-axis) of 5 different seeds at each iteration (x-axis) and the shaded area represents one standard deviation from the mean. (c): Ablation result on SparseHalfCheetah. Each iteration represents 50K time steps for SwimmerGather and 5K time steps for SparseHalfCheetah. }
    \label{fig:mujoco_returns}
\end{figure*}

\begin{figure*}[ht!]
    \centering
    \begin{subfigure}{0.32\textwidth}
        \includegraphics[width=\textwidth]{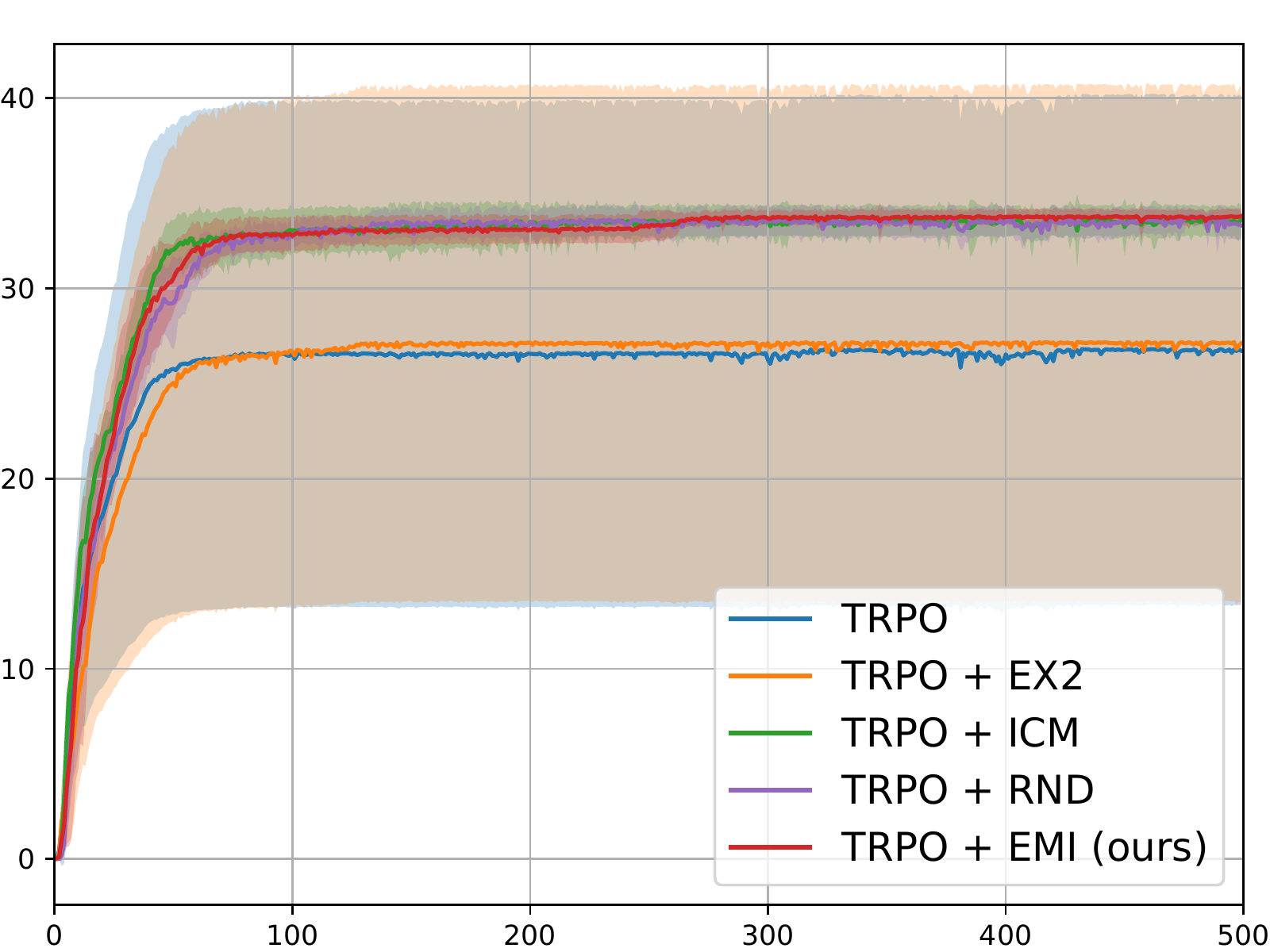}
        \caption{Freeway}
    \end{subfigure}
    \begin{subfigure}{0.32\textwidth}
        \includegraphics[width=\textwidth]{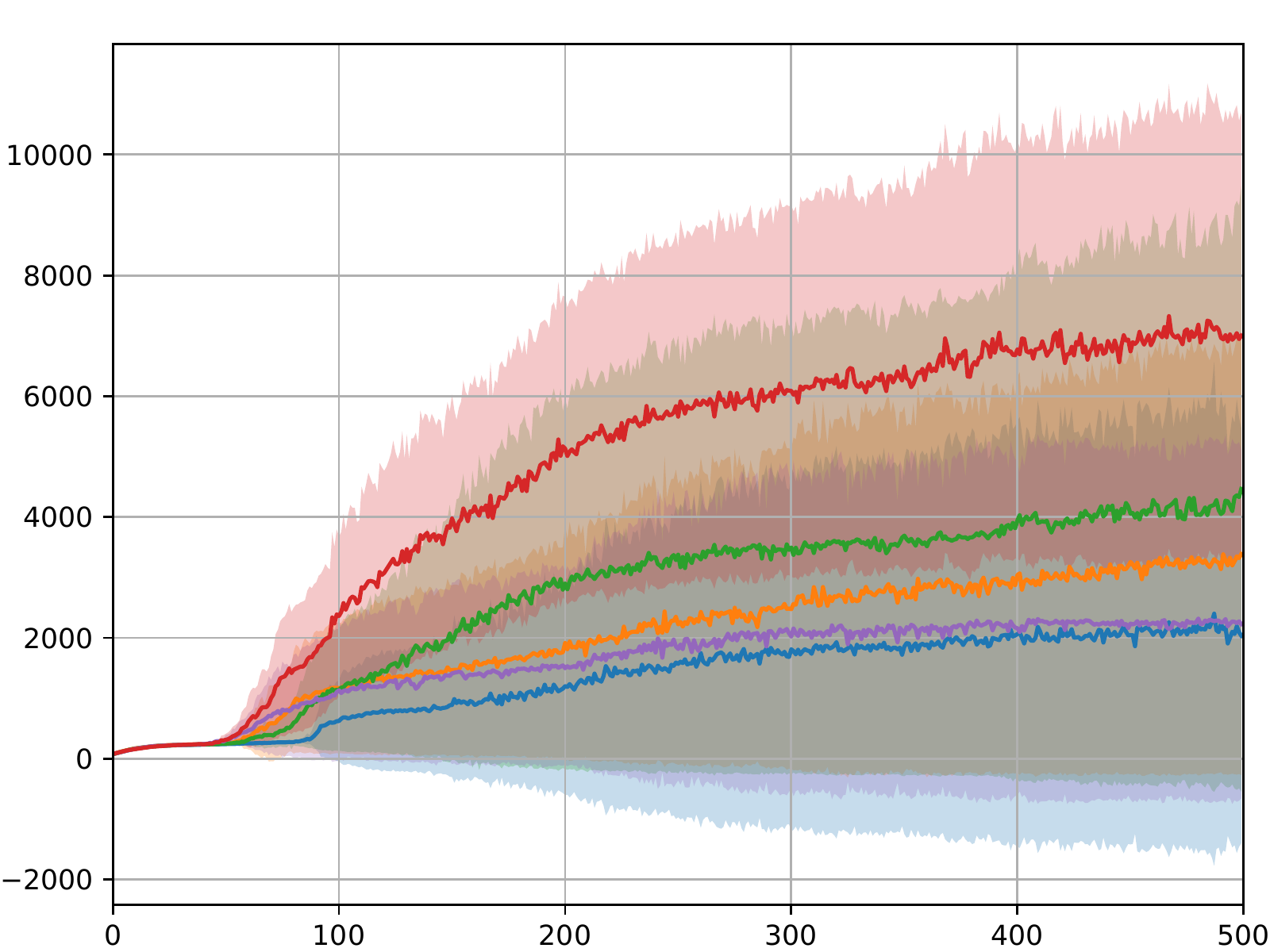}
        \caption{Frostbite}
    \end{subfigure}
    \begin{subfigure}{0.32\textwidth}
        \includegraphics[width=\textwidth]{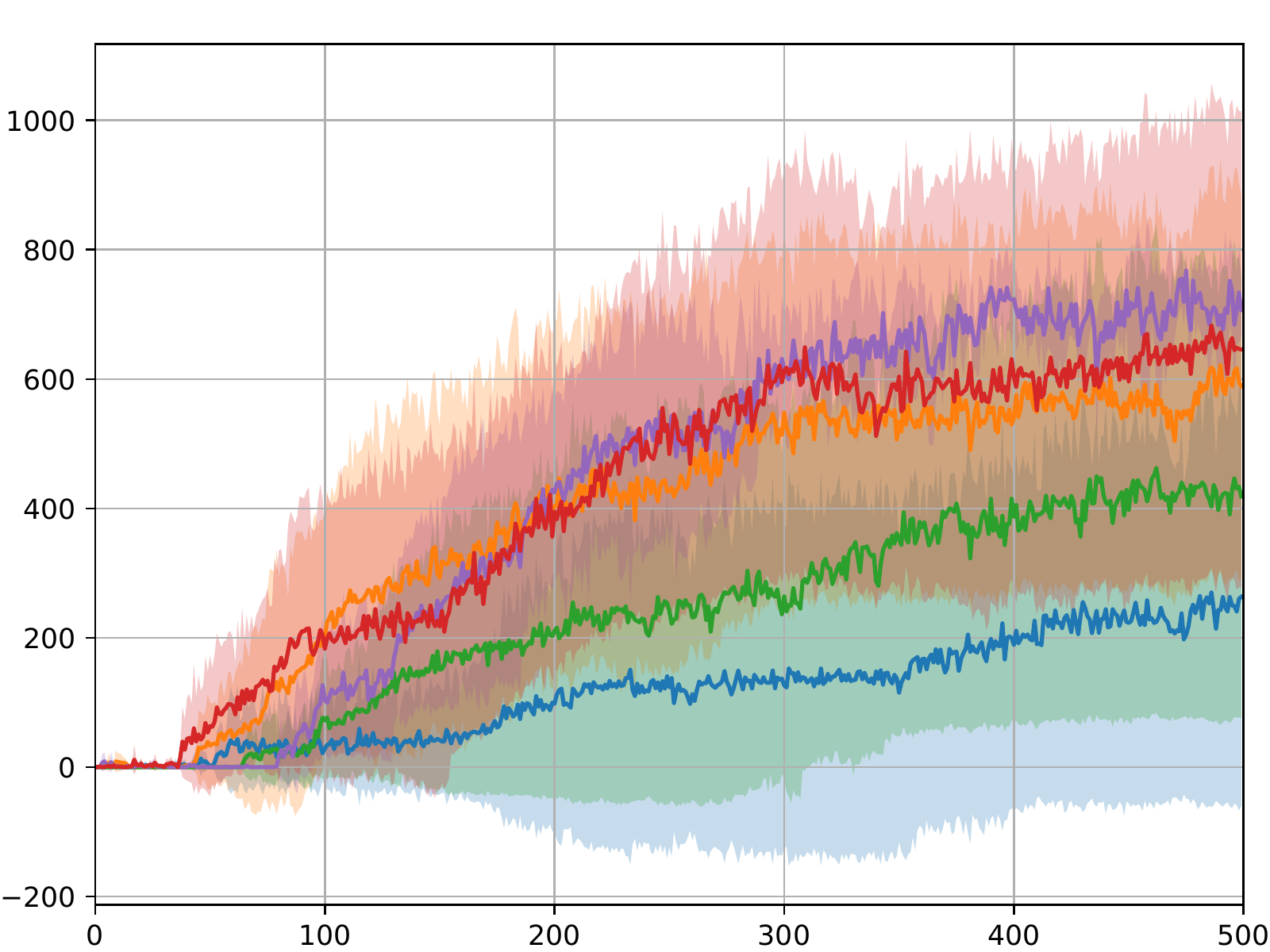}
        \caption{Venture}
    \end{subfigure}
    
    \begin{subfigure}{0.32\textwidth}
        \includegraphics[width=\textwidth]{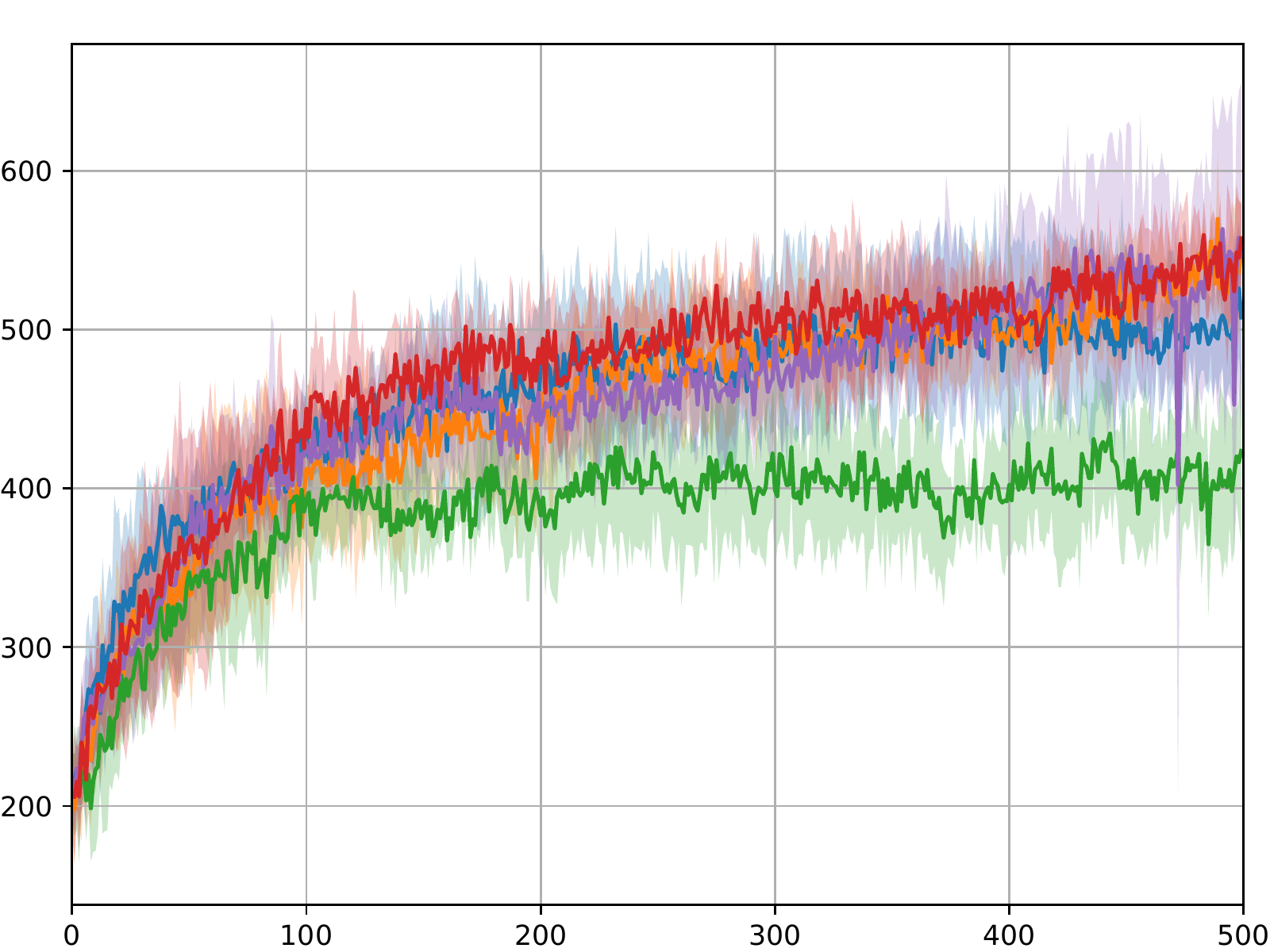}
        \caption{Gravitar}
    \end{subfigure}
    \begin{subfigure}{0.32\textwidth}
        \includegraphics[width=\textwidth]{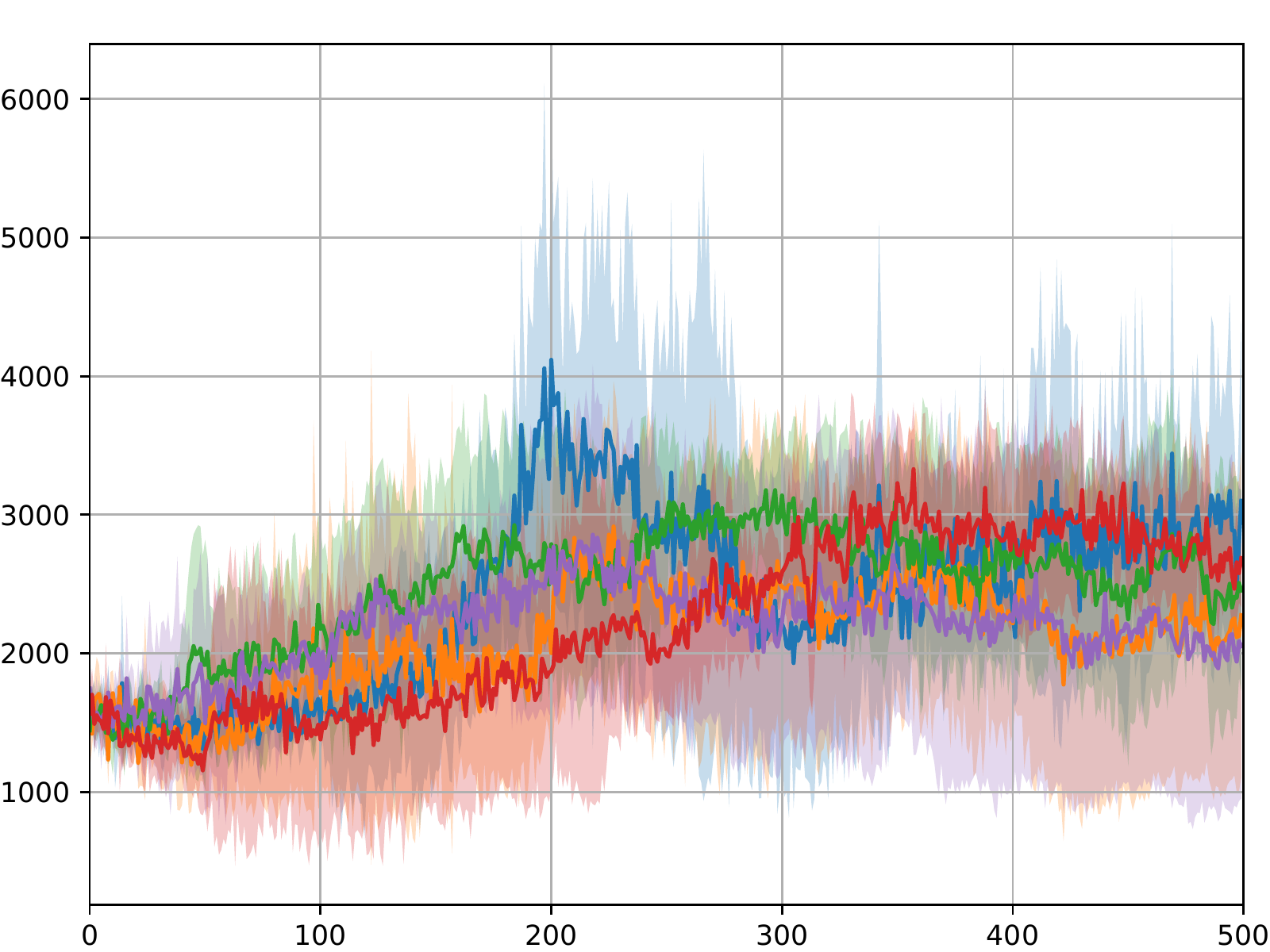}
        \caption{Solaris}
    \end{subfigure}
    \begin{subfigure}{0.32\textwidth}
        \includegraphics[width=\textwidth]{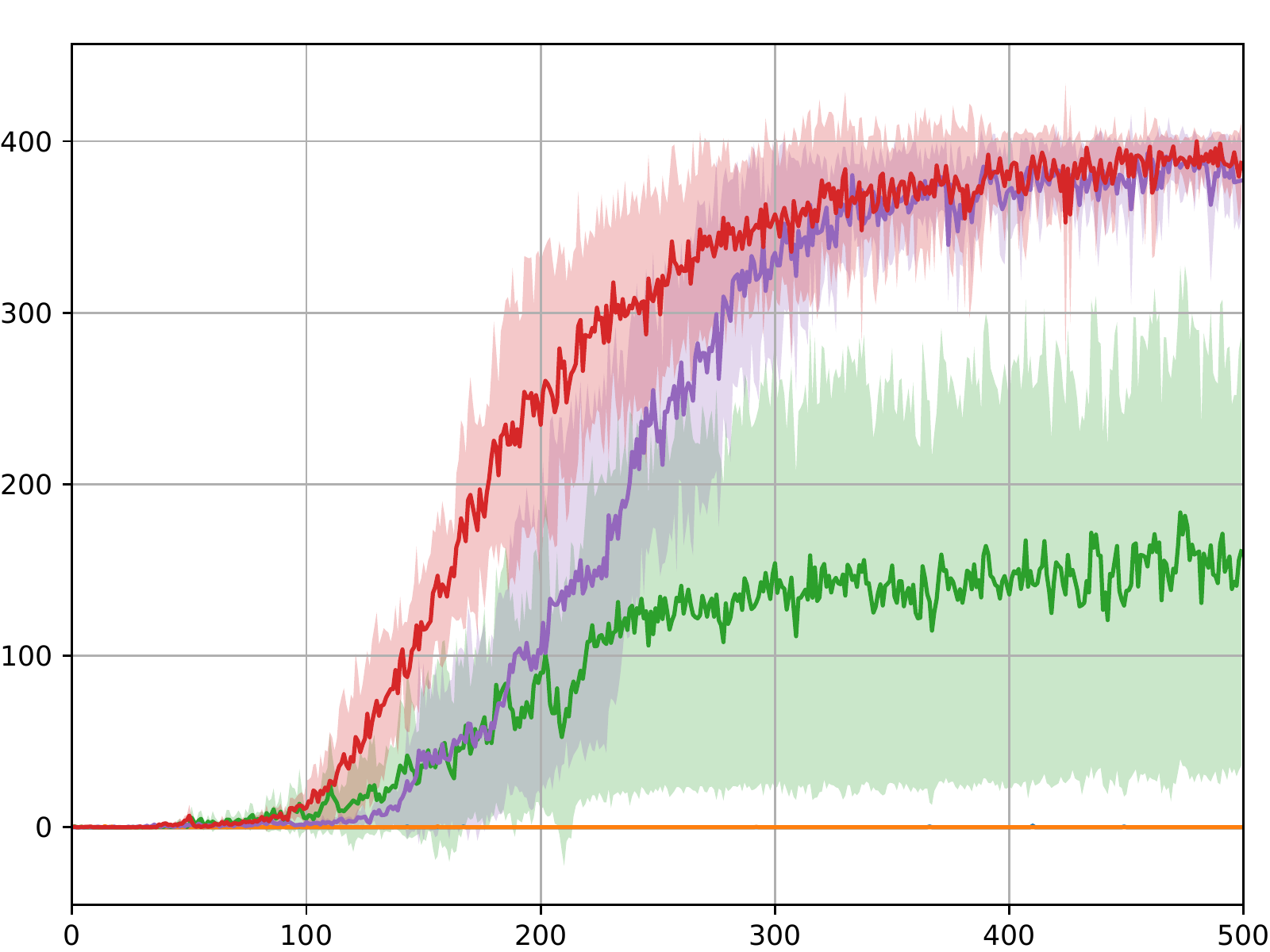}
        \caption{Montezuma's Revenge}
    \end{subfigure}
    \caption{Performance of EMI on sparse reward Atari environments compared to the baseline methods. The solid lines show the mean reward (y-axis) of 5 different seeds at each iteration (x-axis). Each iteration represents 100K time steps.}
    \label{fig:atari_returns}
\vspace{-1em}
\end{figure*}

\subsection{Locomotion tasks with continuous control}
We compare EMI with TRPO \citep{trpo}, EX2 \citep{ex2}, ICM \citep{icm} and RND \citep{rnd} on two challenging locomotion environments: SwimmerGather and SparseHalfCheetah. \Cref{fig:SwimmerGather_returns,fig:SparseHalfCheetah_returns} shows that EMI outperforms all baseline methods on both tasks. \Cref{fig:paths_emb_sparsehalfcheetah} visualizes the scatter plot of the learned state embeddings and an example trajectory for the SparseHalfCheetah experiment. The figure shows that the learned representation successfully preserves the similarity in observation space.

\subsection{Vision-based tasks with discrete control}
For vision-based exploration tasks, our results in \Cref{fig:atari_returns} show that EMI significantly outperforms the TRPO, EX2, ICM baselines on Frostbite and Montezuma's Revenge, and show competitive performance against RND. \Cref{fig:paths_emb_montezuma,fig:paths_emb_frostbite} illustrate our learned state embeddings $\phi$. Since our embedding dimensionality is set to $d=2$, we directly visualize the scatter plot of the embedding representation in 2D. \Cref{fig:paths_emb_montezuma} shows that the embedding space naturally separates state samples into two clusters each of which corresponds to different rooms in Montezuma's revenge. \Cref{fig:paths_emb_frostbite} shows smooth sample transitions along the embedding space in Frostbite where functionally similar states are close together and distinct states are far apart.

\begin{table*}[ht!]
\begin{tabular}{ |p{2.8cm}| >{\centering\arraybackslash}p{1.5cm} | >{\centering\arraybackslash}p{1.5cm}|  >{\centering\arraybackslash}p{1.5cm} | >{\centering\arraybackslash}p{1.5cm} | >{\centering\arraybackslash}p{2cm} | >{\centering\arraybackslash}p{1.5cm} | >{\centering\arraybackslash}p{1.5cm} |}
\hline
& EMI & EX2 & ICM  & RND & AE-SimHash & VIME & TRPO  \\
\hline
SwimmerGather& \textbf{0.438}& 0.200& 0 & 0 & 0.258& 0.196& 0\\
SparseHalfCheetah& \textbf{218.1}& 153.7& 1.4& 3.4 & 0.5& 98.0& 0\\
\hline
Freeway & \textbf{33.8}& 27.1& 33.6& 33.3 & 33.5& -& 26.7\\

Frostbite & \textbf{7002}& 3387& 4465& 2227 & 5214& -& 2034\\

Venture & 646 & 589& 418& \textbf{707} & 445& -& 263\\

Gravitar & \textbf{558}& 550& 424& 546 & 482& -& 508\\

Solaris & 2688& 2276& 2453& 2051 & \textbf{4467}& -& 3101\\

Montezuma & \textbf{387}& 0& 161& 377 & 75& -& 0\\
\hline

 \hline
\end{tabular}
\vspace{-0.7em}
\caption{Mean reward comparison of baseline methods. We compare EMI with EX2 \citep{ex2}, ICM \citep{icm}, RND \citep{rnd}, AE-SimHash \citep{hash_exploration}, VIME \citep{vime}, and TRPO \citep{trpo}. The EMI, EX2, ICM, RND, and TRPO columns show the mean reward of 5 different seeds consistent with the settings in \Cref{fig:mujoco_returns} and \Cref{fig:atari_returns}. The AE-SimHash and VIME columns show the results from the original papers. All methods in the table are implemented based on TRPO policy. The results of MuJoCo experiments are reported at 5M and 100M time steps respectively. The results of Atari experiments are reported at 50M time steps.}
\label{table:compare}
\vspace{-1em}
\end{table*}

\subsection{Ablation study}
We perform an ablation study showing the effect of removing each term in the objective in \Cref{eqn:master_eqn} on SparseHalfCheetah. First, removing the information gain term collapses the embedding space and the agent fails to get any rewards as shown in \Cref{fig:mujoco_ablation}. Also, we observed that adding the model error term (Purple versus Red in the figure) shows drastic performance improvement. We observed that modeling the linear dynamics error helps stabilize the embedding learning process during training. Please refer to supplementary Section 4, 5, and 6 for further analyses.

\subsection{Regularization of embedding distributions}  \label{sec:regularization_target}
\begin{wrapfigure}{r}{0.2\textwidth}
  \vspace{-1.5em}
  \begin{center}
    \includegraphics[width=0.09\textwidth]{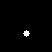}
    \includegraphics[width=0.09\textwidth]{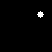}
  \end{center}
  \caption{Example observations from BoxImage. White agent moves inside the black box.}
  \label{fig:boximage_obs_example}
  \vspace{-1em}
\end{wrapfigure}
In order to visually examine the learned embedding representations, we designed a simple image-based 2D environment which we call \textit{BoxImage}. In BoxImage, the agent exists at a position with real-valued coordinates and moves by performing actions in a confined 2D space. Then the agent receives the top-down view of the environment as image states (examples shown in \Cref{fig:boximage_obs_example}). For the implementation details, please refer to supplementary Section 7.

\begin{figure}[ht!]
    \centering
    \captionsetup[subfigure]{labelformat=empty}
    \begin{subfigure}[t]{0.15\textwidth}
        \centering
        \includegraphics[width=\textwidth]{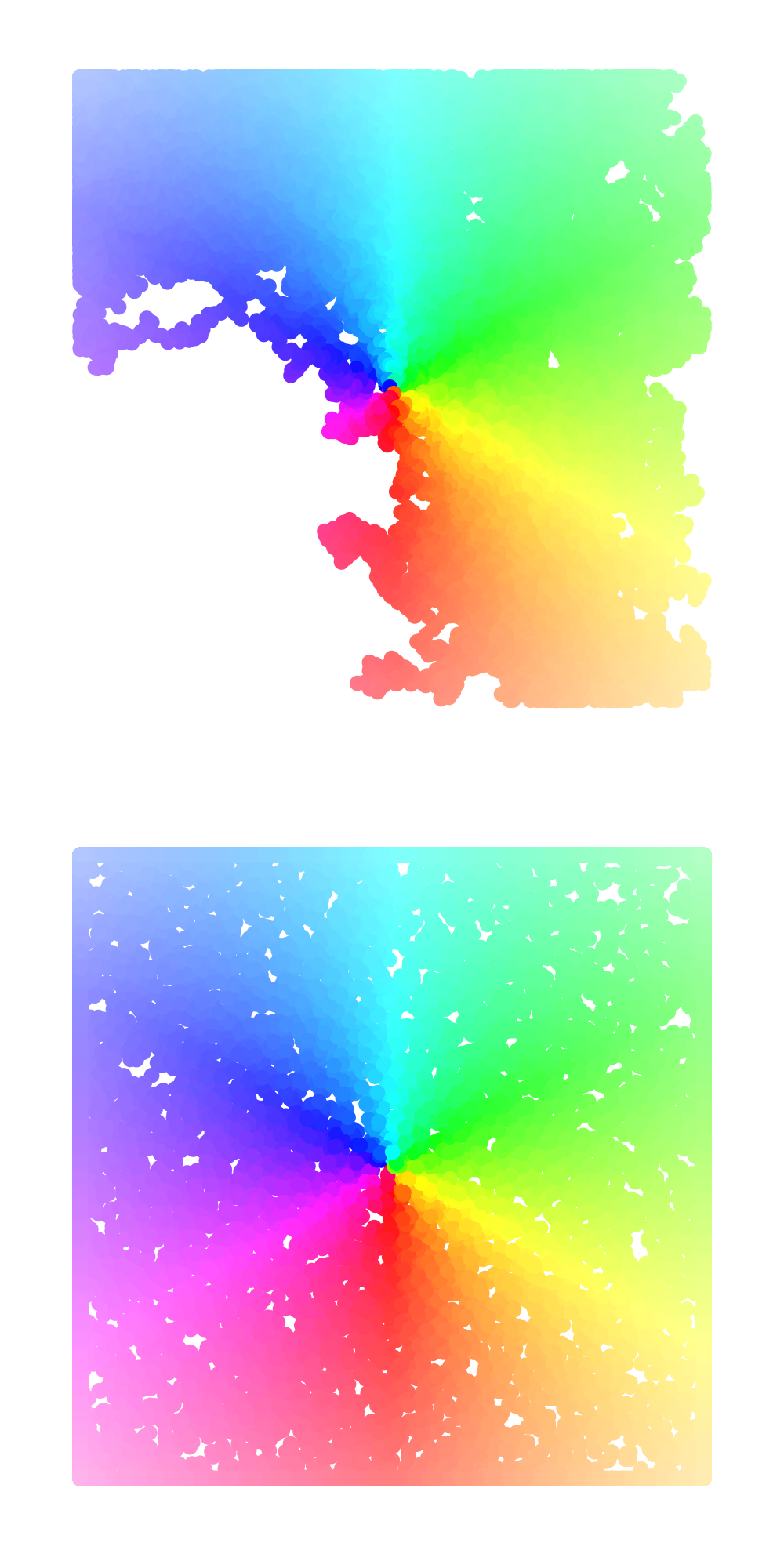}
        \label{fig:boximage_regularization_target_positions_actions}
    \end{subfigure}
    \unskip\ \vrule\
    \begin{subfigure}[t]{0.15\textwidth}
        \centering
        \includegraphics[width=\textwidth]{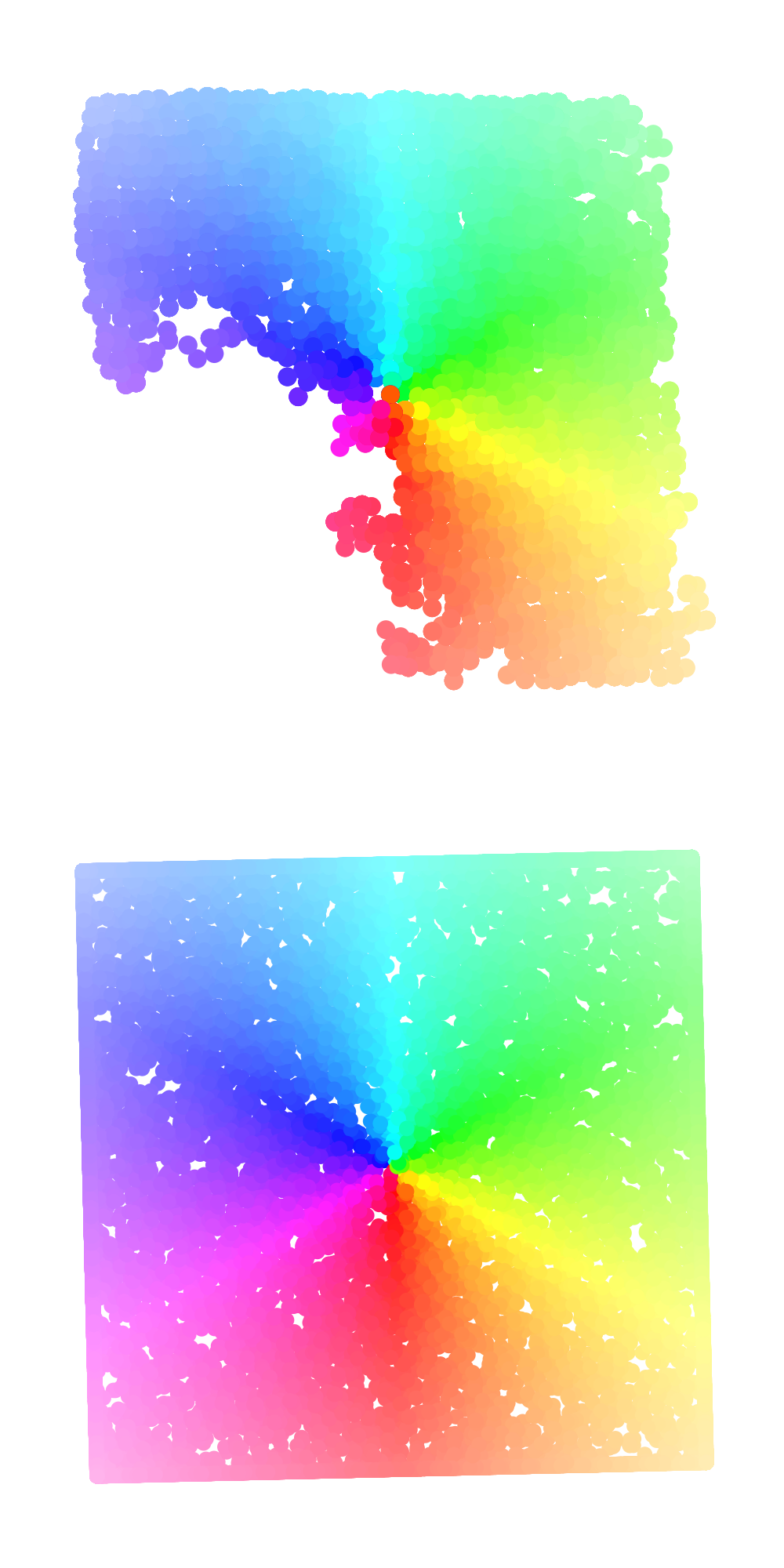}
        \caption{\scriptsize$\mathbf{D_\text{KL}(\mathbb{P}^\pi_\psi \parallel \mathcal{N}(0,I))}$\normalsize}
        \label{fig:boximage_regularization_target_action_prior}
    \end{subfigure}
    \begin{subfigure}[t]{0.15\textwidth}
        \centering
        \includegraphics[width=\textwidth]{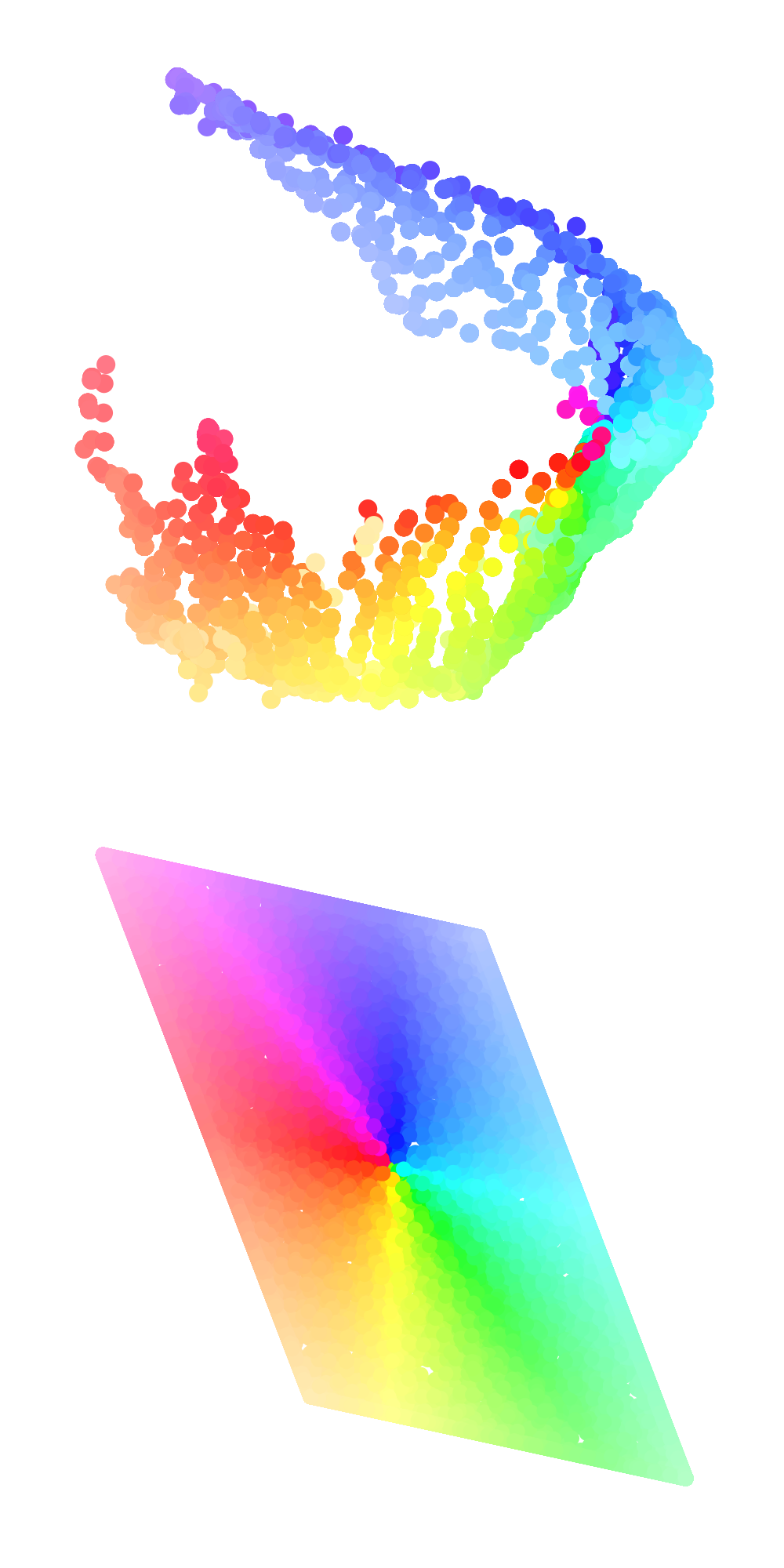}
        \caption{\scriptsize$\mathbf{D_\text{KL}(\mathbb{P}^\pi_\phi \parallel \mathcal{N}(0,I))}$\normalsize}
        \label{fig:boximage_regularization_target_state_prior}
    \end{subfigure}
    \vspace{-0.5em}
    \caption{(Left) Agent's actual 2D positions and actions at the top and bottom respectively. (Center) Learned state and action embeddings when the \emph{action} embedding is regularized. (Right) Learned state and action embeddings when the \emph{state} embedding is regularized.}
    \label{fig:boximage_regularization_target}
    \vspace{-1.7em}
\end{figure}

When the state embedding function $\phi: \reals^{52\times52}\rightarrow \mathbb{R}^2$, and the action embedding function $\psi: \reals^2 \rightarrow \reals^2$ are trained with the regularization on the \emph{action} embedding distribution with $D_\text{KL}(\mathbb{P}^\pi_\psi \parallel \mathcal{N}(0,I))$, the learned embedding representations successfully represent the distributions of the agent's 2D positions and actions, as shown in \Cref{fig:boximage_regularization_target}. On the other hand, employing the regularization on the \emph{state} embedding distribution with $D_\text{KL}(\mathbb{P}^\pi_\phi \parallel \mathcal{N}(0,I))$ results in severe degradation in the embedding quality, mainly due to the skewness of the  state sample distribution.


\vspace{-0.5em}
\section{Conclusion}
We presented EMI, a practical exploration method that does not rely on the direct generation of high dimensional observations and instead extracts the predictive signal that can be used for exploration within a compact representation space. Our results on challenging robotic locomotion tasks with continuous actions and high dimensional image-based games with sparse rewards show that our approach transfers to a wide range of tasks. As future work, we would like to explore utilizing the learned linear dynamic model for optimal planning in the embedding representation space. In particular, we would like to investigate how an optimal trajectory from a state to a given goal in the embedding space under the linear representation topology translates to the optimal trajectory in the observation space under complex dynamical systems.

\vspace{-1em}
\section*{Acknowledgements}\vspace{-0.5em}
This work was partially supported by Samsung Advanced Institute of Technology and Institute for Information \& Communications Technology Planning \& Evaluation (IITP) grant funded by the Korea government (MSIT) (No.2019-0-01367, BabyMind). Hyun Oh Song is the corresponding author.

\bibliography{main}

\begin{thebibliography}{36}
\providecommand{\natexlab}[1]{#1}
\providecommand{\url}[1]{\texttt{#1}}
\expandafter\ifx\csname urlstyle\endcsname\relax
  \providecommand{\doi}[1]{doi: #1}\else
  \providecommand{\doi}{doi: \begingroup \urlstyle{rm}\Url}\fi

\bibitem[Belghazi et~al.(2018)Belghazi, Rajeswar, Baratin, Hjelm, and
  Courville]{mine}
Belghazi, I., Rajeswar, S., Baratin, A., Hjelm, R.~D., and Courville, A.
\newblock Mutual information neural estimation.
\newblock In \emph{International Conference on Machine Learning}, volume 2018,
  2018.

\bibitem[Bellemare et~al.(2014)Bellemare, Veness, and Talvitie]{bellemare14}
Bellemare, M., Veness, J., and Talvitie, E.
\newblock Skip context tree switching.
\newblock In \emph{International Conference on Machine Learning}, pp.\
  1458--1466, 2014.

\bibitem[Bellemare et~al.(2016)Bellemare, Srinivasan, Ostrovski, Schaul,
  Saxton, and Munos]{bellemare16}
Bellemare, M., Srinivasan, S., Ostrovski, G., Schaul, T., Saxton, D., and
  Munos, R.
\newblock Unifying count-based exploration and intrinsic motivation.
\newblock In \emph{Advances in Neural Information Processing Systems}, pp.\
  1471--1479, 2016.

\bibitem[Bellemare et~al.(2013)Bellemare, Naddaf, Veness, and Bowling]{ale}
Bellemare, M.~G., Naddaf, Y., Veness, J., and Bowling, M.
\newblock The arcade learning environment: An evaluation platform for general
  agents.
\newblock \emph{Journal of Artificial Intelligence Research}, 47:\penalty0
  253--279, 2013.

\bibitem[Bengio et~al.(2017)Bengio, Thomas, Pineau, Precup, and
  Bengio]{bengio2017}
Bengio, E., Thomas, V., Pineau, J., Precup, D., and Bengio, Y.
\newblock Independently controllable features.
\newblock In \emph{Multidisciplinary Conference on Reinforcement Learning and
  Decision Making}, 2017.

\bibitem[Burda et~al.(2018)Burda, Edwards, Storkey, and Klimov]{rnd}
Burda, Y., Edwards, H., Storkey, A., and Klimov, O.
\newblock Exploration by random network distillation.
\newblock \emph{arXiv preprint arXiv:1810.12894}, 2018.

\bibitem[Cand{\`e}s et~al.(2011)Cand{\`e}s, Li, Ma, and Wright]{robustpca}
Cand{\`e}s, E.~J., Li, X., Ma, Y., and Wright, J.
\newblock Robust principal component analysis?
\newblock \emph{Journal of the ACM (JACM)}, 58\penalty0 (3):\penalty0 11, 2011.

\bibitem[Chen et~al.(2016)Chen, Duan, Houthooft, Schulman, Sutskever, and
  Abbeel]{infogan}
Chen, X., Duan, Y., Houthooft, R., Schulman, J., Sutskever, I., and Abbeel, P.
\newblock Infogan: Interpretable representation learning by information
  maximizing generative adversarial nets.
\newblock In \emph{Advances in neural information processing systems}, pp.\
  2172--2180, 2016.

\bibitem[Colas et~al.(2018)Colas, Sigaud, and Oudeyer]{colas2018gep}
Colas, C., Sigaud, O., and Oudeyer, P.-Y.
\newblock Gep-pg: Decoupling exploration and exploitation in deep reinforcement
  learning algorithms.
\newblock \emph{arXiv preprint arXiv:1802.05054}, 2018.

\bibitem[Donsker \& Varadhan(1983)Donsker and Varadhan]{donsker1983}
Donsker, M.~D. and Varadhan, S.~S.
\newblock Asymptotic evaluation of certain markov process expectations for
  large time. iv.
\newblock \emph{Communications on Pure and Applied Mathematics}, 36\penalty0
  (2):\penalty0 183--212, 1983.

\bibitem[Duan et~al.(2016)Duan, Chen, Houthooft, Schulman, and Abbeel]{rllab}
Duan, Y., Chen, X., Houthooft, R., Schulman, J., and Abbeel, P.
\newblock Benchmarking deep reinforcement learning for continuous control.
\newblock In \emph{International Conference on Machine Learning}, pp.\
  1329--1338, 2016.

\bibitem[Fu et~al.(2017)Fu, Co-Reyes, and Levine]{ex2}
Fu, J., Co-Reyes, J., and Levine, S.
\newblock Ex2: Exploration with exemplar models for deep reinforcement
  learning.
\newblock In \emph{Advances in Neural Information Processing Systems}, pp.\
  2577--2587, 2017.

\bibitem[Goodfellow et~al.(2014)Goodfellow, Pouget-Abadie, Mirza, Xu,
  Warde-Farley, Ozair, Courville, and Bengio]{gan}
Goodfellow, I., Pouget-Abadie, J., Mirza, M., Xu, B., Warde-Farley, D., Ozair,
  S., Courville, A., and Bengio, Y.
\newblock Generative adversarial nets.
\newblock In \emph{Advances in neural information processing systems}, pp.\
  2672--2680, 2014.

\bibitem[Hjelm et~al.(2018)Hjelm, Fedorov, Lavoie-Marchildon, Grewal,
  Trischler, and Bengio]{dim}
Hjelm, R.~D., Fedorov, A., Lavoie-Marchildon, S., Grewal, K., Trischler, A.,
  and Bengio, Y.
\newblock Learning deep representations by mutual information estimation and
  maximization.
\newblock \emph{arXiv preprint arXiv:1808.06670}, 2018.

\bibitem[Houthooft et~al.(2016)Houthooft, Chen, Duan, Schulman, De~Turck, and
  Abbeel]{vime}
Houthooft, R., Chen, X., Duan, Y., Schulman, J., De~Turck, F., and Abbeel, P.
\newblock Vime: Variational information maximizing exploration.
\newblock In \emph{Advances in Neural Information Processing Systems}, pp.\
  1109--1117, 2016.

\bibitem[Kingma \& Ba(2015)Kingma and Ba]{adam}
Kingma, D.~P. and Ba, J.~L.
\newblock Adam: A method for stochastic optimization.
\newblock In \emph{Proceedings of the 3rd International Conference on Learning
  Representations (ICLR)}, 2015.

\bibitem[Kingma \& Welling(2013)Kingma and Welling]{vae}
Kingma, D.~P. and Welling, M.
\newblock Auto-encoding variational bayes.
\newblock \emph{arXiv preprint arXiv:1312.6114}, 2013.

\bibitem[Kohonen(1983)]{kohonen1983}
Kohonen, T.
\newblock Representation of information in spatial maps which are produced by
  self-organization.
\newblock In \emph{Synergetics of the Brain}, pp.\  264--273. Springer, 1983.

\bibitem[Kohonen \& Somervuo(1998)Kohonen and Somervuo]{kohonen1998}
Kohonen, T. and Somervuo, P.
\newblock Self-organizing maps of symbol strings.
\newblock \emph{Neurocomputing}, 21\penalty0 (1-3):\penalty0 19--30, 1998.

\bibitem[Mnih et~al.(2015)Mnih, Kavukcuoglu, Silver, Rusu, Veness, Bellemare,
  Graves, Riedmiller, Fidjeland, Ostrovski, et~al.]{dqn_nature}
Mnih, V., Kavukcuoglu, K., Silver, D., Rusu, A.~A., Veness, J., Bellemare,
  M.~G., Graves, A., Riedmiller, M., Fidjeland, A.~K., Ostrovski, G., et~al.
\newblock Human-level control through deep reinforcement learning.
\newblock \emph{Nature}, 518\penalty0 (7540):\penalty0 529, 2015.

\bibitem[Mnih et~al.(2016)Mnih, Badia, Mirza, Graves, Lillicrap, Harley,
  Silver, and Kavukcuoglu]{a3c}
Mnih, V., Badia, A.~P., Mirza, M., Graves, A., Lillicrap, T., Harley, T.,
  Silver, D., and Kavukcuoglu, K.
\newblock Asynchronous methods for deep reinforcement learning.
\newblock In \emph{International conference on machine learning}, pp.\
  1928--1937, 2016.

\bibitem[Mohamed \& Rezende(2015)Mohamed and Rezende]{mohamed}
Mohamed, S. and Rezende, D.~J.
\newblock Variational information maximisation for intrinsically motivated
  reinforcement learning.
\newblock In \emph{Advances in neural information processing systems}, pp.\
  2125--2133, 2015.

\bibitem[Nachum et~al.(2018)Nachum, Gu, Lee, and Levine]{nachum2018near}
Nachum, O., Gu, S., Lee, H., and Levine, S.
\newblock Near-optimal representation learning for hierarchical reinforcement
  learning.
\newblock \emph{arXiv preprint arXiv:1810.01257}, 2018.

\bibitem[Ng et~al.(1999)Ng, Harada, and Russell]{andrewng_rewardshaping}
Ng, A.~Y., Harada, D., and Russell, S.
\newblock Policy invariance under reward transformations: Theory and
  application to reward shaping.
\newblock In \emph{ICML}, volume~99, pp.\  278--287, 1999.

\bibitem[Nowozin et~al.(2016)Nowozin, Cseke, and Tomioka]{fgan}
Nowozin, S., Cseke, B., and Tomioka, R.
\newblock f-gan: Training generative neural samplers using variational
  divergence minimization.
\newblock In \emph{Advances in Neural Information Processing Systems}, pp.\
  271--279, 2016.

\bibitem[Oh et~al.(2015)Oh, Guo, Lee, Lewis, and Singh]{action_conditional}
Oh, J., Guo, X., Lee, H., Lewis, R.~L., and Singh, S.
\newblock Action-conditional video prediction using deep networks in atari
  games.
\newblock In \emph{Advances in neural information processing systems}, pp.\
  2863--2871, 2015.

\bibitem[Oord et~al.(2018)Oord, Li, and Vinyals]{cpc}
Oord, A. v.~d., Li, Y., and Vinyals, O.
\newblock Representation learning with contrastive predictive coding.
\newblock \emph{arXiv preprint arXiv:1807.03748}, 2018.

\bibitem[Ostrovski et~al.(2017)Ostrovski, Bellemare, Oord, and
  Munos]{ostrovski17}
Ostrovski, G., Bellemare, M.~G., Oord, A. v.~d., and Munos, R.
\newblock Count-based exploration with neural density models.
\newblock \emph{arXiv preprint arXiv:1703.01310}, 2017.

\bibitem[Pathak et~al.(2017)Pathak, Agrawal, Efros, and Darrell]{icm}
Pathak, D., Agrawal, P., Efros, A.~A., and Darrell, T.
\newblock Curiosity-driven exploration by self-supervised prediction.
\newblock In \emph{International Conference on Machine Learning}, volume 2017,
  2017.

\bibitem[Schulman et~al.(2015)Schulman, Levine, Abbeel, Jordan, and
  Moritz]{trpo}
Schulman, J., Levine, S., Abbeel, P., Jordan, M., and Moritz, P.
\newblock Trust region policy optimization.
\newblock In \emph{International Conference on Machine Learning}, volume 2015,
  2015.

\bibitem[Schulman et~al.(2017)Schulman, Wolski, Dhariwal, Radford, and
  Klimov]{ppo}
Schulman, J., Wolski, F., Dhariwal, P., Radford, A., and Klimov, O.
\newblock Proximal policy optimization algorithms.
\newblock \emph{arXiv preprint arXiv:1707.06347}, 2017.

\bibitem[Stadie et~al.(2015)Stadie, Levine, and Abbeel]{stadie}
Stadie, B.~C., Levine, S., and Abbeel, P.
\newblock Incentivizing exploration in reinforcement learning with deep
  predictive models.
\newblock \emph{arXiv preprint arXiv:1507.00814}, 2015.

\bibitem[Tang et~al.(2017)Tang, Houthooft, Foote, Stooke, Chen, Duan, Schulman,
  DeTurck, and Abbeel]{hash_exploration}
Tang, H., Houthooft, R., Foote, D., Stooke, A., Chen, X., Duan, Y., Schulman,
  J., DeTurck, F., and Abbeel, P.
\newblock \# exploration: A study of count-based exploration for deep
  reinforcement learning.
\newblock In \emph{Advances in Neural Information Processing Systems}, pp.\
  2753--2762, 2017.

\bibitem[Thomas et~al.(2017{\natexlab{a}})Thomas, Bengio, Fedus, Pondard,
  Beaudoin, Larochelle, Pineau, Precup, and Bengio]{thomas2017}
Thomas, V., Bengio, E., Fedus, W., Pondard, J., Beaudoin, P., Larochelle, H.,
  Pineau, J., Precup, D., and Bengio, Y.
\newblock Disentangling the independently controllable factors of variation by
  interacting with the world.
\newblock In \emph{NIPS2017 Workshop on Learning Disentangled Representations:
  from Perception to Control}, pp.\  1--5, 2017{\natexlab{a}}.

\bibitem[Thomas et~al.(2017{\natexlab{b}})Thomas, Pondard, Bengio, Sarfati,
  Beaudoin, Meurs, Pineau, Precup, and Bengio]{icf}
Thomas, V., Pondard, J., Bengio, E., Sarfati, M., Beaudoin, P., Meurs, M.-J.,
  Pineau, J., Precup, D., and Bengio, Y.
\newblock Independently controllable factors.
\newblock \emph{arXiv preprint arXiv:1708.01289}, 2017{\natexlab{b}}.

\bibitem[van~den Oord et~al.(2016)van~den Oord, Kalchbrenner, Espeholt,
  Vinyals, Graves, et~al.]{vandenoord16}
van~den Oord, A., Kalchbrenner, N., Espeholt, L., Vinyals, O., Graves, A.,
  et~al.
\newblock Conditional image generation with pixelcnn decoders.
\newblock In \emph{Advances in Neural Information Processing Systems}, pp.\
  4790--4798, 2016.

\end{thebibliography}


\begin{thebibliography}{3}
\providecommand{\natexlab}[1]{#1}
\providecommand{\url}[1]{\texttt{#1}}
\expandafter\ifx\csname urlstyle\endcsname\relax
  \providecommand{\doi}[1]{doi: #1}\else
  \providecommand{\doi}{doi: \begingroup \urlstyle{rm}\Url}\fi

\bibitem[Colas et~al.(2018)Colas, Sigaud, and Oudeyer]{colas2018gep}
Colas, C., Sigaud, O., and Oudeyer, P.-Y.
\newblock Gep-pg: Decoupling exploration and exploitation in deep reinforcement
  learning algorithms.
\newblock \emph{arXiv preprint arXiv:1802.05054}, 2018.

\bibitem[Ng et~al.(1999)Ng, Harada, and Russell]{andrewng_rewardshaping}
Ng, A.~Y., Harada, D., and Russell, S.
\newblock Policy invariance under reward transformations: Theory and
  application to reward shaping.
\newblock In \emph{ICML}, volume~99, pp.\  278--287, 1999.

\bibitem[Oh et~al.(2015)Oh, Guo, Lee, Lewis, and Singh]{action_conditional}
Oh, J., Guo, X., Lee, H., Lewis, R.~L., and Singh, S.
\newblock Action-conditional video prediction using deep networks in atari
  games.
\newblock In \emph{Advances in neural information processing systems}, pp.\
  2863--2871, 2015.

\end{thebibliography}
\bibliographystyle{icml2019}

\clearpage

\appendix
\twocolumn[
\icmltitle{Supplementary}
]
\renewcommand{\thesection}{\arabic{section}}
\section{Experiment Hyperparameters} \label{sec:appendix_hparams}

In all of our experiments, we use Adam optimizer with the learning rate of 0.001 and a minibatch size of 512 for 3 epochs to optimize embedding networks. In each iteration, we train the embedding networks. The embedding dimensionality is set to $d=2$ and the intrinsic reward coefficient is set as 0.001 in all environments. \Cref{table:mujoco_hp} and \Cref{table:atari_hp} give the detailed information of the remaining hyperparameters.

\begin{table*}
\centering
\begin{tabular}{ |p{3cm}| >{\centering\arraybackslash}p{5cm} | >{\centering\arraybackslash}p{5cm}|  }
\hline
Environments & SwimmerGather & SparseHalfCheetah\\
\hline
\hline
TRPO method & \multicolumn{2}{c|}{Single Path} \\
\hline
TRPO step size & \multicolumn{2}{c|}{0.01} \\
\hline
TRPO batch size  & 50k & 5k\\
\hline
Policy network & \multicolumn{2}{c|}{ A 2-layer FC with (64, 32) hidden units (tanh)  } \\
\hline
Baseline network & A 32 hidden units FC (ReLU) & Linear baseline \\
\hline
$\lambda_{\text{error}}$ & 0.001 & 5 \\
\hline
$\lambda_{\text{info}}$ & \multicolumn{2}{c|}{1} \\
\hline
$\phi$ network & \multicolumn{2}{c|}{Same structure as policy network} \\
\hline
$\psi$ network & \multicolumn{2}{c|}{A 64 hidden units FC (ReLU)} \\
\hline
Information network & \multicolumn{2}{c|}{A 2-layer FC with (64, 64) hidden units (ReLU) }\\
\hline 
Error network & \multicolumn{2}{c|}{\makecell{State input passes the same network structure as policy network. \\
Concat layer concatenates state output and action. \\ A 256 units FC (ReLU) }} \\
\hline
Max path length & \multicolumn{2}{c|}{500} \\
\hline
Discount factor & \multicolumn{2}{c|}{0.995} \\

 \hline
\end{tabular}
\caption{Hyperparameters for MuJoCo experiments.}
\label{table:mujoco_hp}
\end{table*}

\begin{table*}[ht!]
\centering
\begin{tabular}{ |p{3cm} | >{\centering\arraybackslash}p{10.5cm}| }
 \hline
 Environments & Freeway, Frostbite, Venture, Montezuma's Revenge, Gravitar, Solaris\\
 \hline
 \hline
TRPO method & Single Path \\
 \hline
TRPO step size & 0.01 \\
 \hline
TRPO batch size  & 100k \\
\hline
Policy network & 2 convolutional layers (16 8x8 filters of stride 4, 32 4x4 filters of stride 2), followed by a 256 hidden units FC (ReLU) \\
\hline
Baseline network & Same structure as policy network \\
\hline
$\phi$ network & Same structure as policy network \\
\hline
$\psi$ network & A 64 hidden units FC (ReLU) \\
\hline
$\lambda_{\text{error}}$ & $100$ \\
\hline
$\lambda_{\text{info}}$ & 0.1 \\
\hline
Information network & A 2-layer FC with (64, 64) hidden units (ReLU) \\
\hline 
Error network & \makecell{State input passes the same network structure as policy network. \\
Concat layer concatenates state output and action. \\ A 256 units FC (ReLU) } \\
\hline
Max path length & 4500 \\
\hline
Discount factor & 0.995 \\
\hline
\end{tabular}
\caption{Hyperparameters for Atari experiments.}
\label{table:atari_hp}
\end{table*}

\section{Different intrinsic reward formulation} \label{section:emi-d}
We evaluate the performance under another intrinsic reward function. Apart from prediction error formulation in our main paper, we also consider the relative difference in the novelty of state representations, based on the distance in the embedding space similar to \cite{action_conditional} as shown in \Cref{eqn:ir_div}.

\begin{align}
\label{eqn:ir_div}
&r_d(s_t, a_t, s_t') = g(s_t) - g(s_t'), \\
&\text{\ where }~~ g(s) = \frac{1}{n} \sum_{i=1}^n \exp{\left( -\frac{\|\phi(s) - \phi(s_i)\|^2}{2 \sigma^2} \right)} \nonumber
\end{align}

The relative difference makes sure the intrinsic reward diminishes to zero \citep{andrewng_rewardshaping} once the agent has sufficiently explored the state space. We label EMI using this diversity based intrinsic reward \Cref{eqn:ir_div} as EMI-D.


\vspace{-0.5em}
\begin{figure}[ht!]
    \centering
    \begin{subfigure}{0.35\textwidth}
        \includegraphics[width=\textwidth]{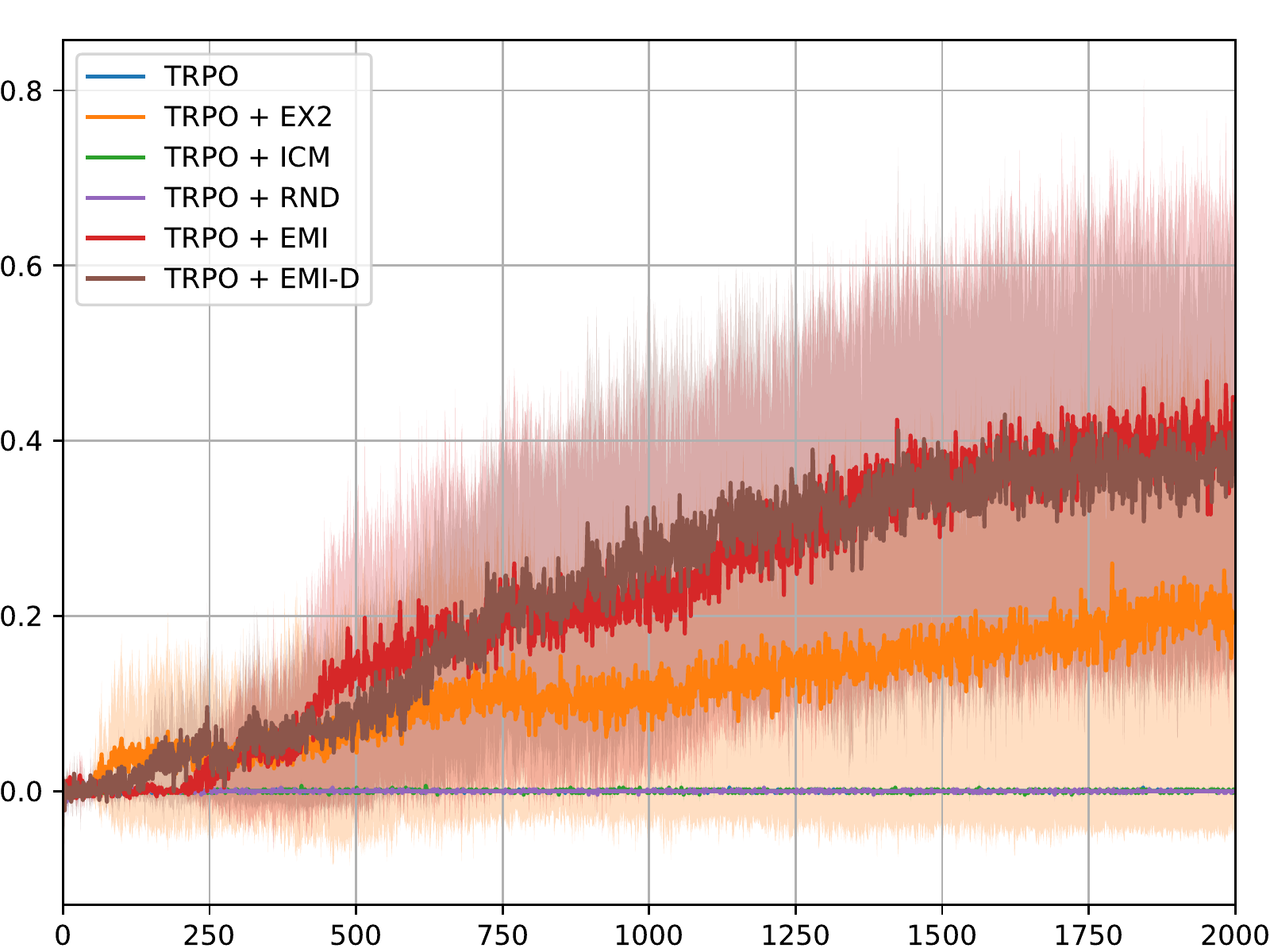}
        \caption{SwimmerGather}
    \end{subfigure}
    \begin{subfigure}{0.35\textwidth}
        \includegraphics[width=\textwidth]{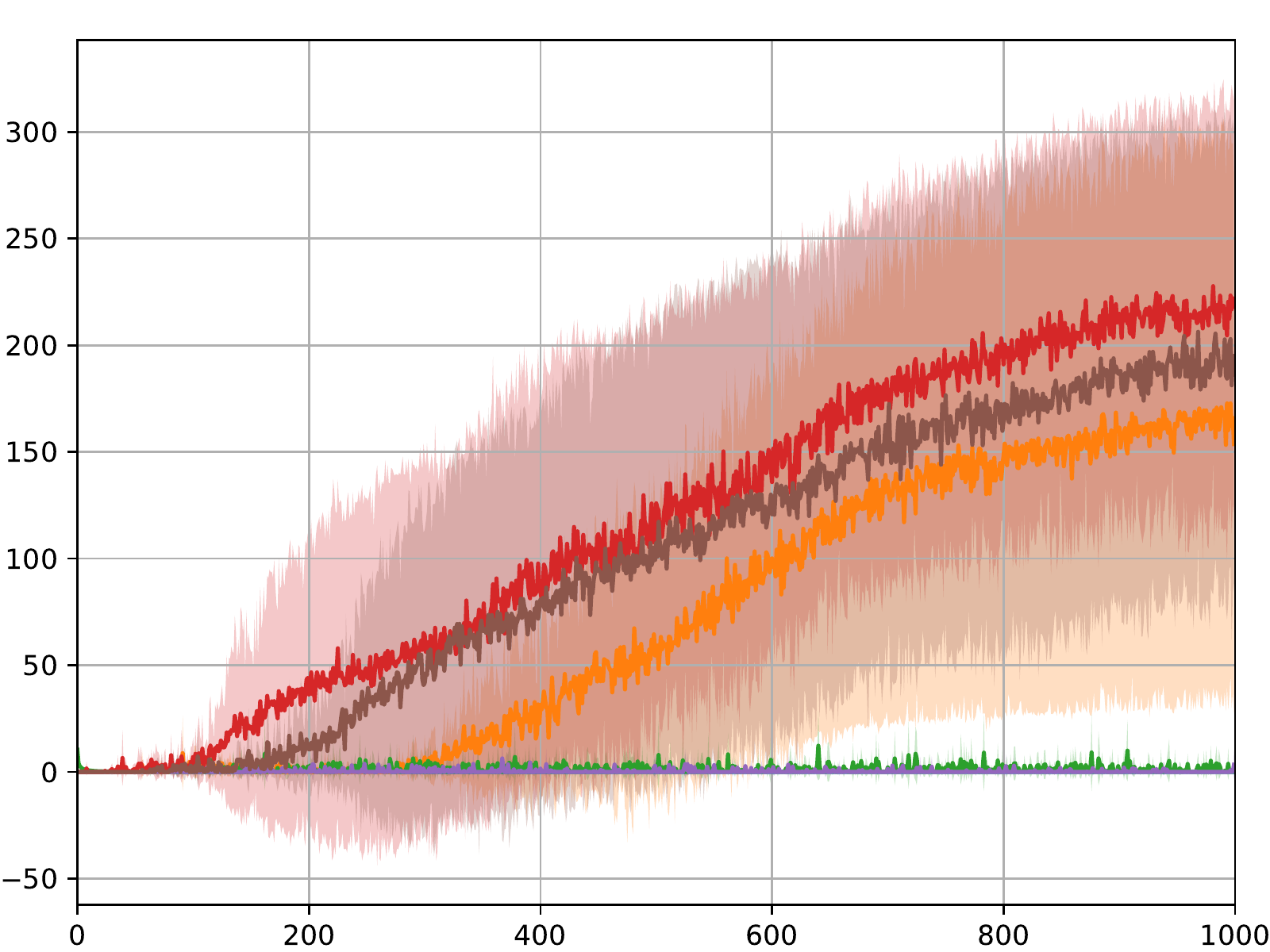}
        \caption{SparseHalfCheetah}
    \end{subfigure}
     \vspace{-0.2cm}
    \caption{Performance of EMI and EMI-D on locomotion tasks with sparse rewards compared to the baseline methods. The solid lines show the mean reward (y-axis) of 5 different seeds at each iteration (x-axis).}
    \label{fig:mujoco_returns_all}
    \vspace{-0.1cm}
\end{figure}

\begin{figure}[ht!]
    \centering
    \includegraphics[width=0.35\textwidth]{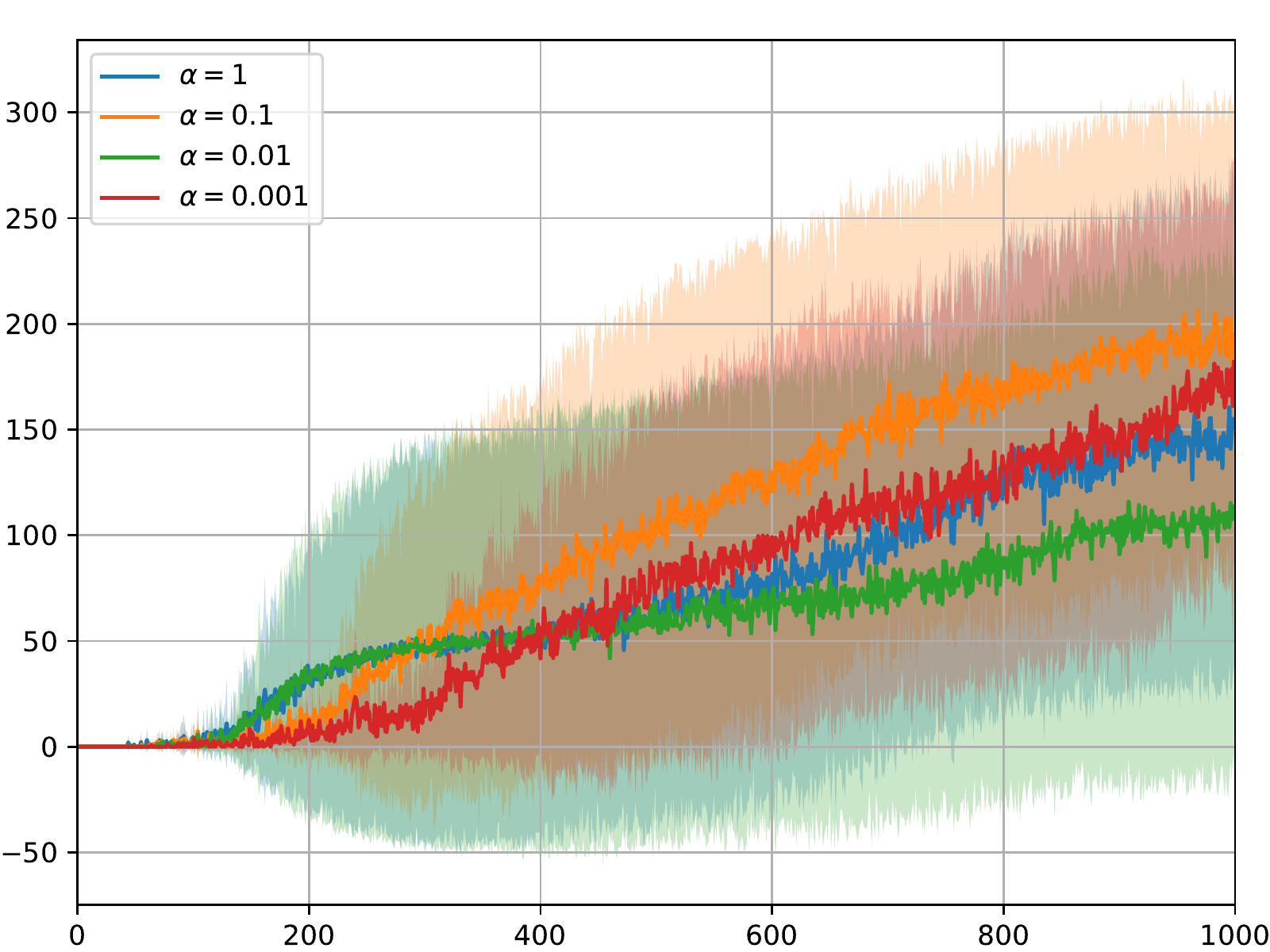}
    \caption{Study of intrinsic reward coefficient $\alpha$ in EMI-D on SparseHalfCheetah environment.}
    \label{fig:ablate_ir}
    \vspace{-1em}
\end{figure}

\begin{figure*}[ht!]
    \centering
    \begin{subfigure}{0.32\textwidth}
        \includegraphics[width=\textwidth]{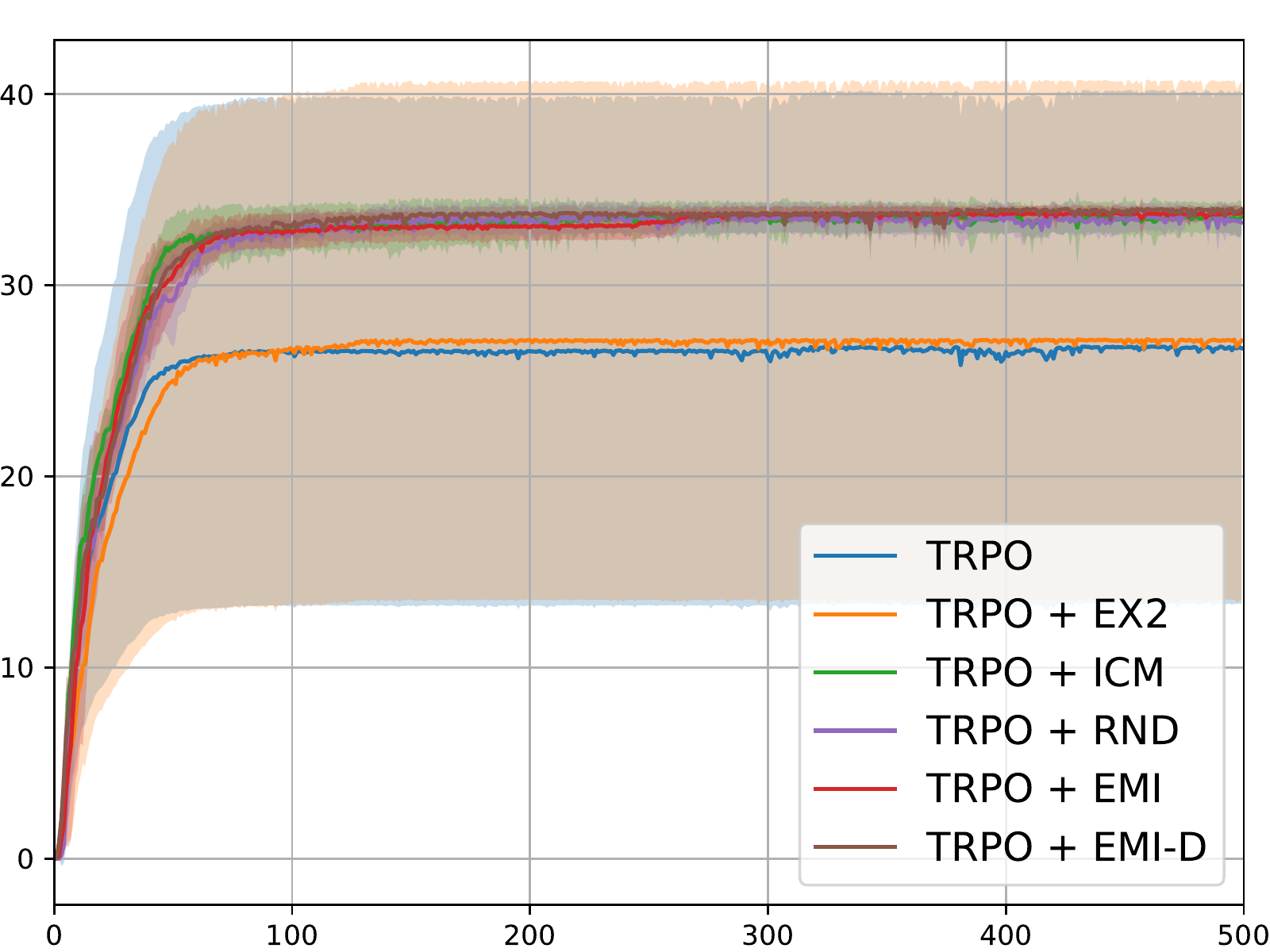}
        \caption{Freeway}
    \end{subfigure}
    \begin{subfigure}{0.32\textwidth}
        \includegraphics[width=\textwidth]{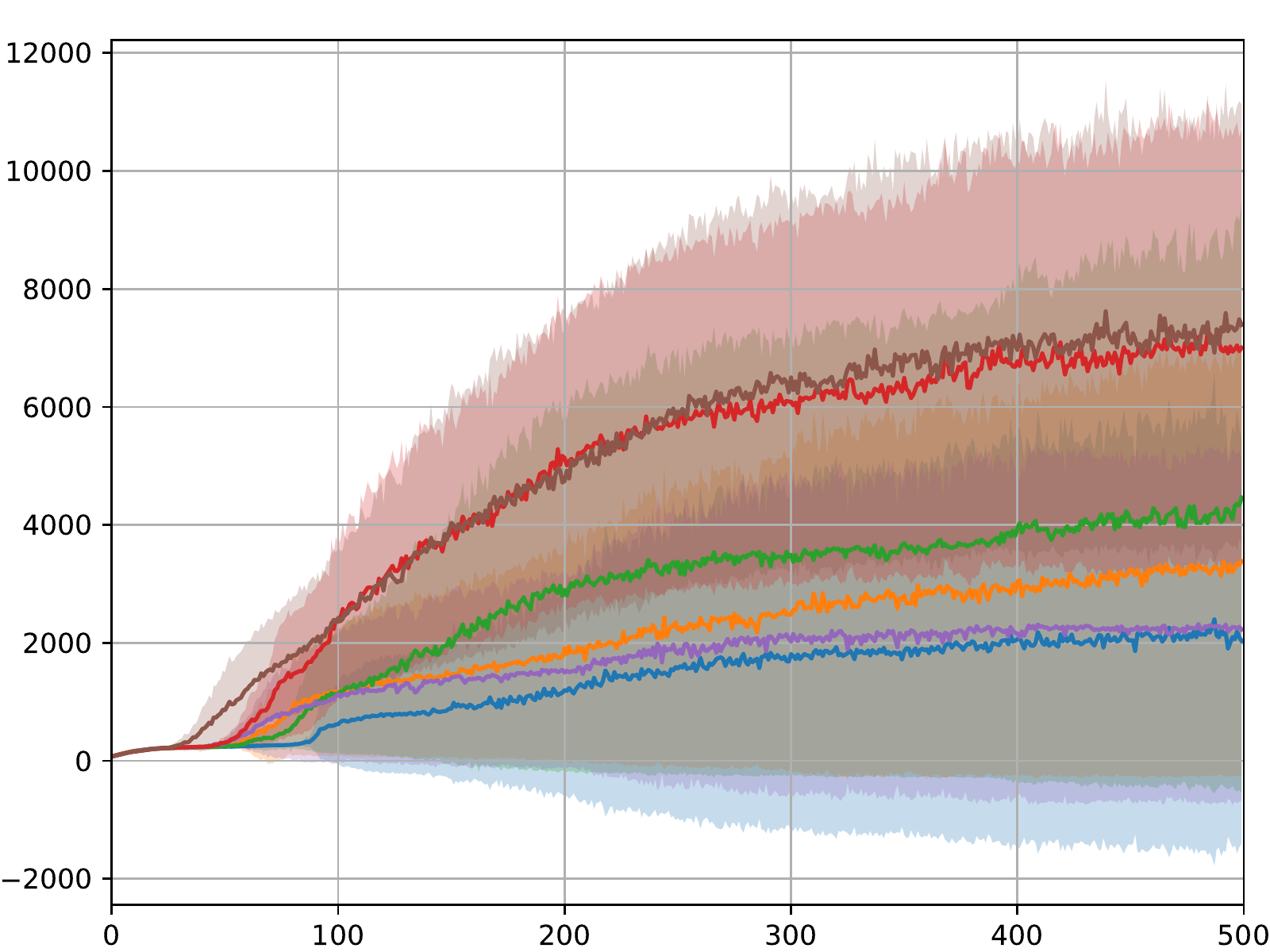}
        \caption{Frostbite}
    \end{subfigure}
    \begin{subfigure}{0.32\textwidth}
        \includegraphics[width=\textwidth]{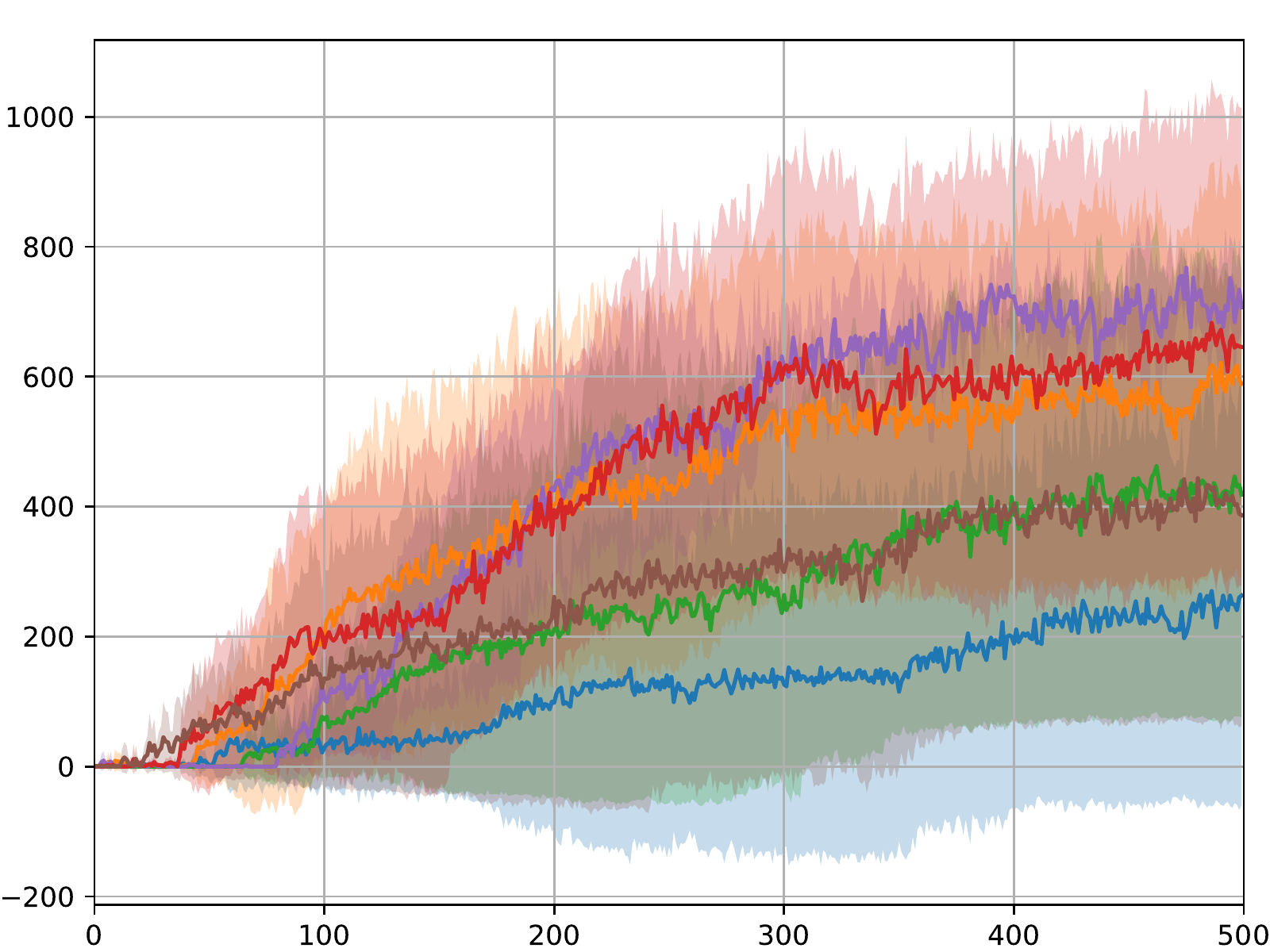}
        \caption{Venture}
    \end{subfigure}
    
    \begin{subfigure}{0.32\textwidth}
        \includegraphics[width=\textwidth]{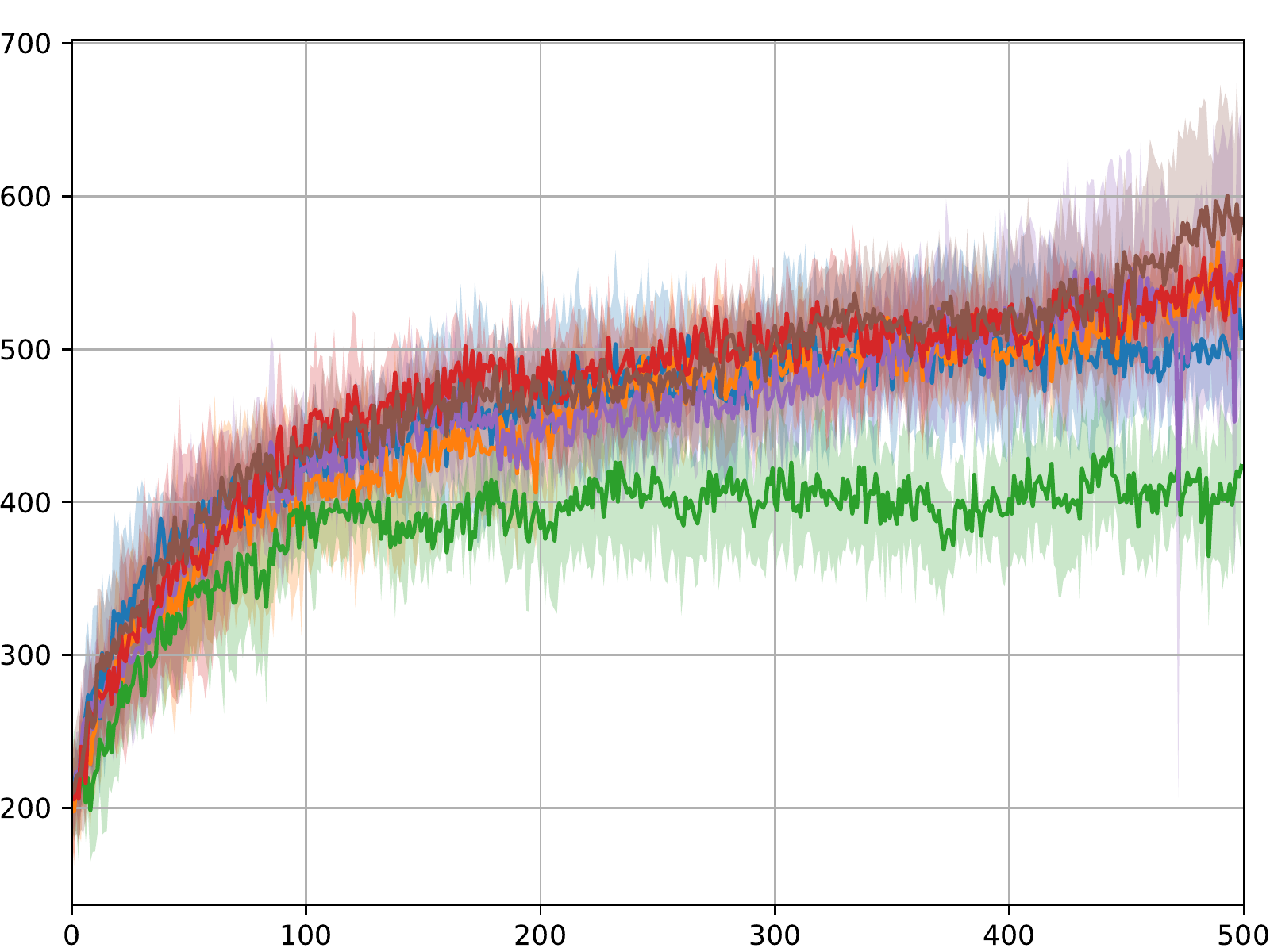}
        \caption{Gravitar}
    \end{subfigure}
    \begin{subfigure}{0.32\textwidth}
        \includegraphics[width=\textwidth]{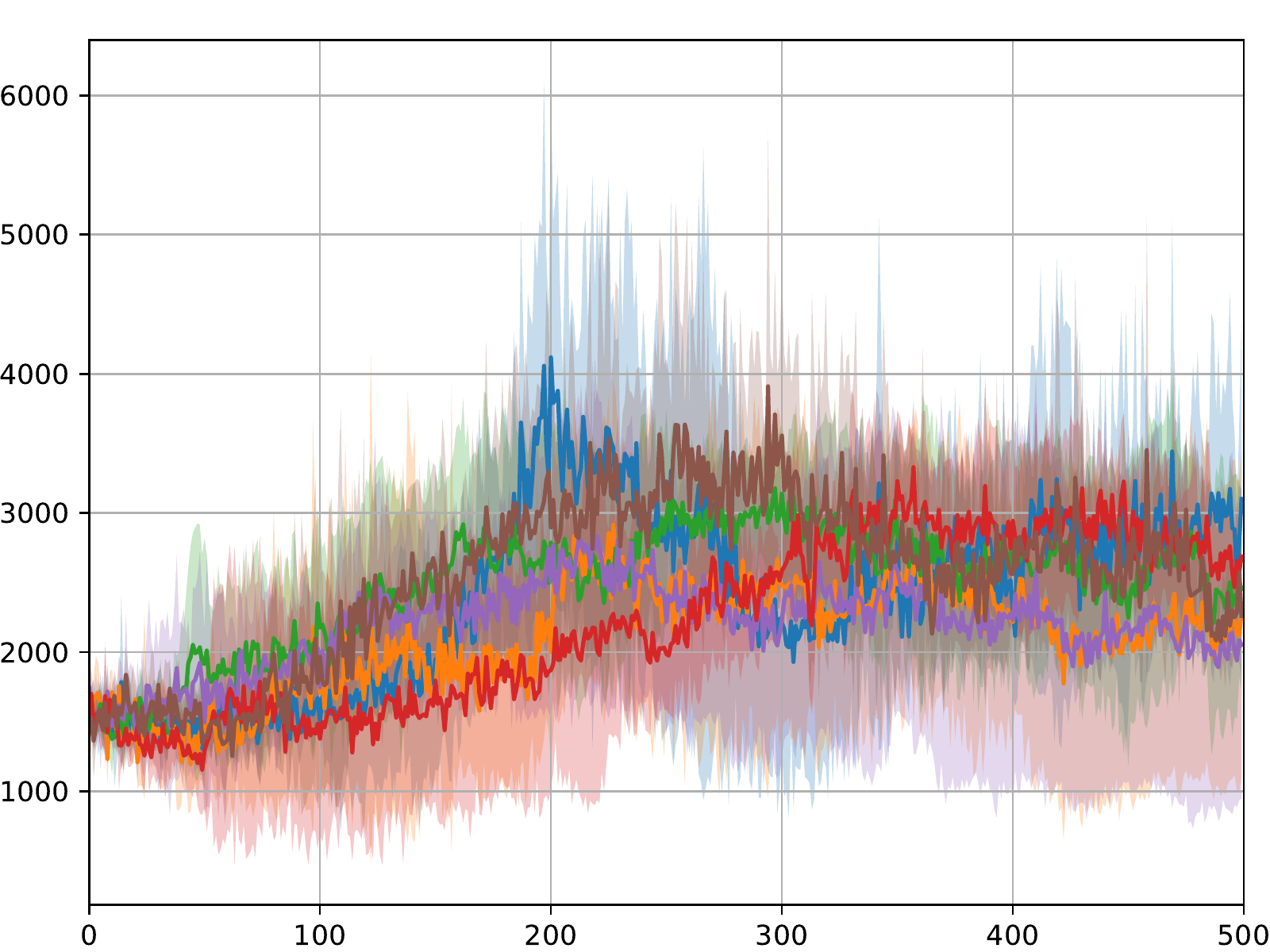}
        \caption{Solaris}
    \end{subfigure}
    \begin{subfigure}{0.32\textwidth}
        \includegraphics[width=\textwidth]{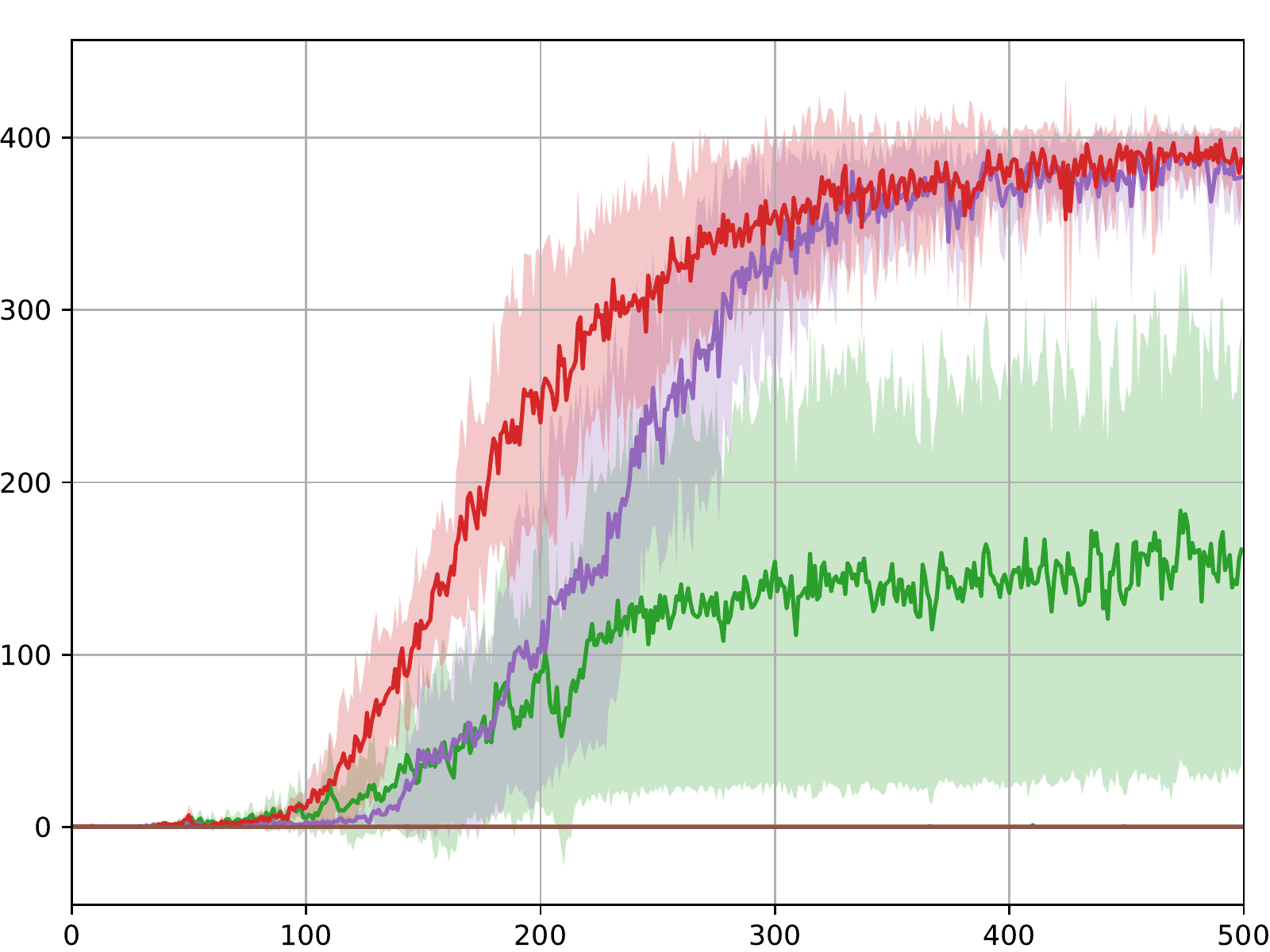}
        \caption{Montezuma's Revenge}
    \end{subfigure}
    \vspace{-0.7em}
    \caption{Performance of EMI and EMI-D on sparse reward Atari environments compared to the baseline methods. The solid lines show the mean reward (y-axis) of 5 different seeds at each iteration (x-axis).}
    \label{fig:atari_returns_all}
\vspace{-0.3em}
\end{figure*}

\Cref{fig:mujoco_returns_all} and \Cref{fig:atari_returns_all} show performance of EMI-D compared to EMI and the baseline exploration methods on MuJoCo and Atari domains respectively. The results show comparable performance in most environments with respect to EMI. In EMI-D, we set $\lambda_\text{info}=0.05$, $\lambda_\text{error}=10000$ and apply action embedding regularization for MuJoCo experiments. For Atari experiments, we use the same hyperparameters as in EMI.

For reward augmentation, EMI-D uses intrinsic reward $r_d$ and then learns from $r = r_{env} + \alpha r_d$. \Cref{fig:ablate_ir} shows the impact of $\alpha$ in EMI-D. Although $\alpha=0.1$ gives the best performance, other choices also give comparable performance.


\begin{figure*}[ht!]
    \centering
    \begin{subfigure}{0.32\textwidth}
        \includegraphics[width=\textwidth]{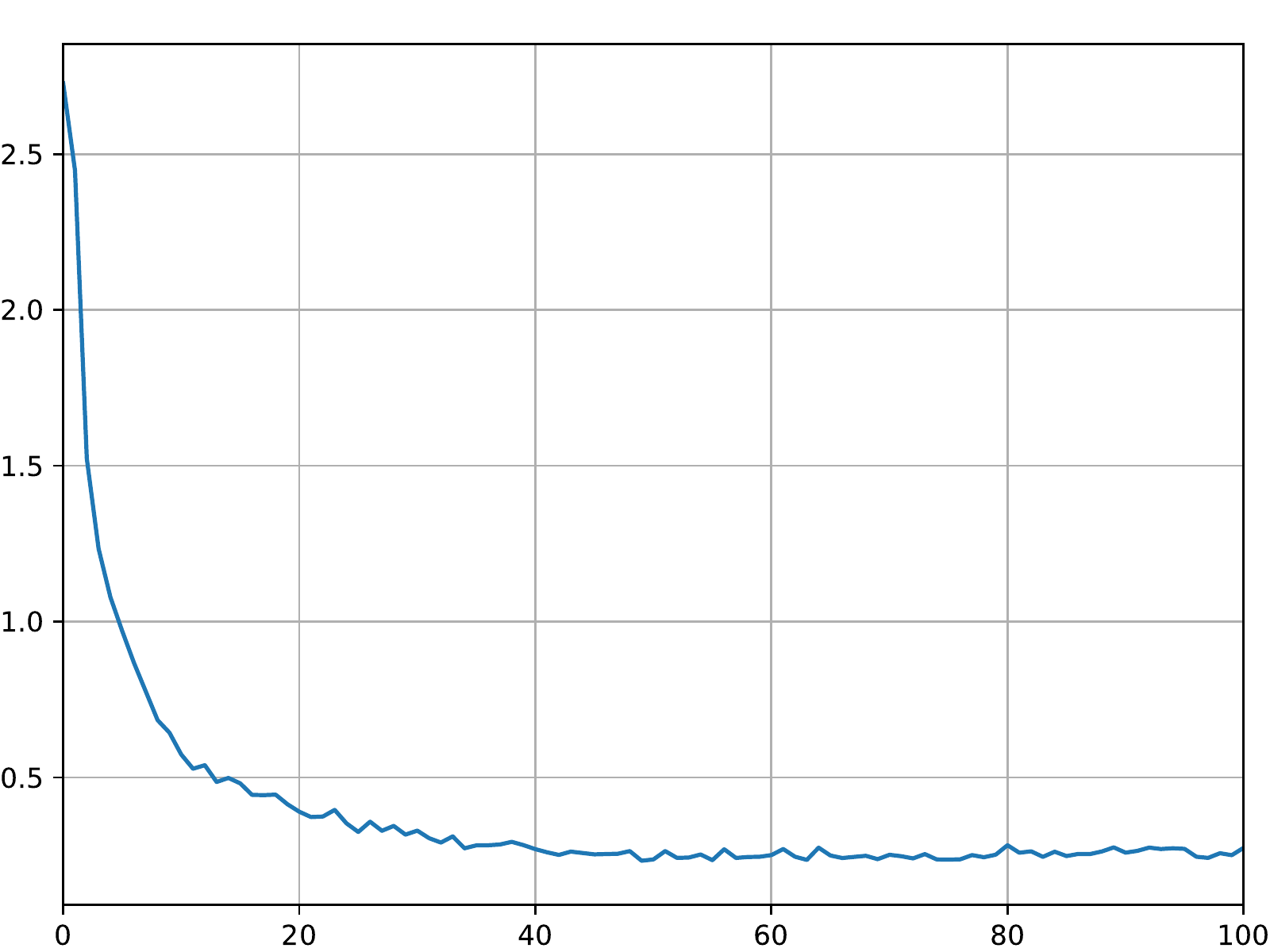}
        \caption{Mutual information loss}
    \end{subfigure}
    \begin{subfigure}{0.32\textwidth}
        \includegraphics[width=\textwidth]{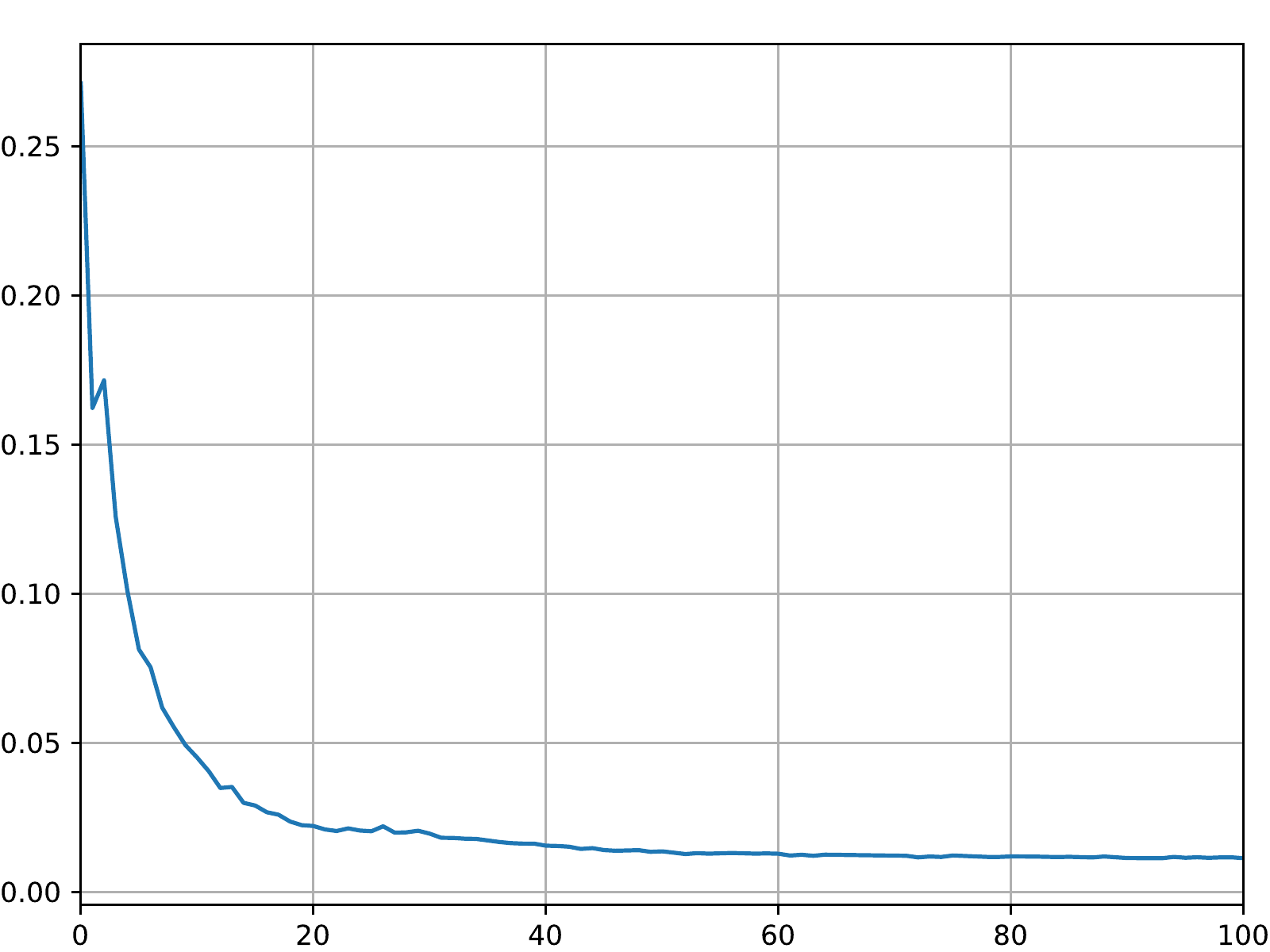}
        \caption{Linear dynamics loss}
    \end{subfigure}
    \begin{subfigure}{0.32\textwidth}
        \includegraphics[width=\textwidth]{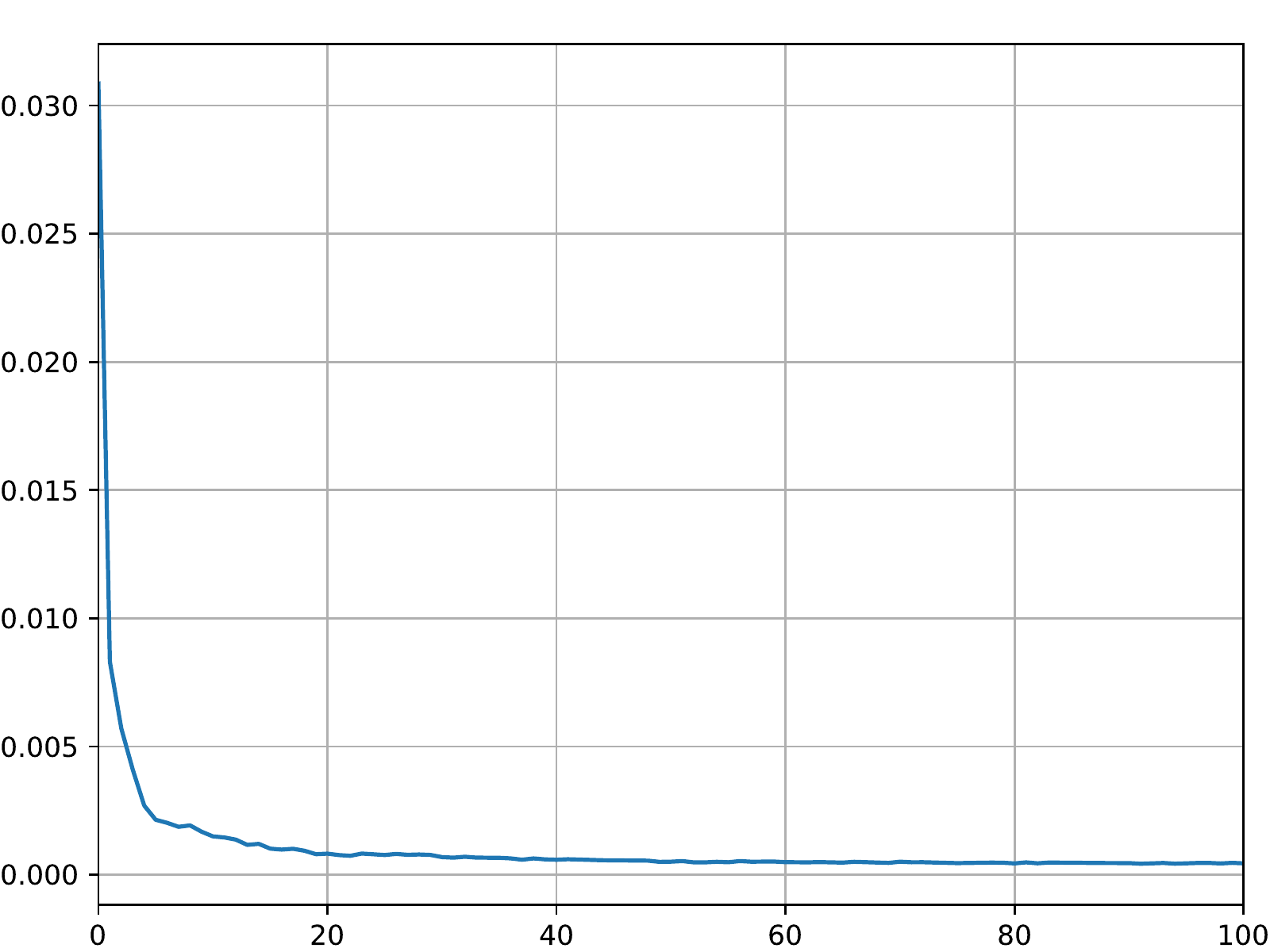}
        \caption{Model error loss}
    \end{subfigure}
    \vspace{-0.7em}
    \caption{Convergence of loss term values (y-axis) across the iterations (x-axis) in SparseHalfCheetah.}
    \label{fig:loss_plots}
\vspace{-0.3em}
\end{figure*}

\section{Computation of the mutual information term}
Given a minibatch $\{(s_{t_l}, a_{t_l}, s_{t_l}') \}_{l=1}^m$, we can construct the following inputs.
\vspace{-3em}
\begin{align*}
    D &= \Big\{\Big(\phi(s_{t_l}), \psi(a_{t_l}), \phi(s_{t_l}')\Big)\Big\}_{l=1}^{\floor{\frac{m}{2}}} \\
    D_s &= \Big\{\Big(\phi(s_{t_l}), \psi(a_{t_l}), \phi\left(s_{t_{l+\floor{\frac{m}{2}}}}'\right)\Big)\Big\}_{l=1}^{\floor{\frac{m}{2}}} \\
    D_a &= \Big\{\left(\phi(s_{t_l}), \psi\left(a_{t_{l+\floor{\frac{m}{2}}}}\right), \phi(s_{t_l}')\right)\Big\}_{l=1}^{\floor{\frac{m}{2}}}
\end{align*}
Then the mutual information term $\mathcal{L}_\text{info}$ in Equation (7) from the main text, is computed as follows.
\small
\begin{align*}
\mathcal{L}_\text{info} =& \inf_{\omega_S \in \Omega_S} \Big[ \mathbb{E}_{d \in D} ~\text{sp} \left(-T_{\omega_S}(d)\right) + \mathbb{E}_{d_s \in D_s} ~\text{sp} \left(T_{\omega_S}(d_s)\right) \Big] \\
+& \inf_{\omega_A \in \Omega_A} \Big[ \mathbb{E}_{d \in D} ~\text{sp} \left(-T_{\omega_A}(d)\right)  + \mathbb{E}_{d_a \in D_a} ~\text{sp} \left(T_{\omega_A}(d_a)\right) \Big]
\end{align*}
\normalsize

\section{Experimental evaluation of the error model} \label{sec:appendix_reconciler}
\begin{figure*}[ht!]
    \centering
    \includegraphics[width=\textwidth]{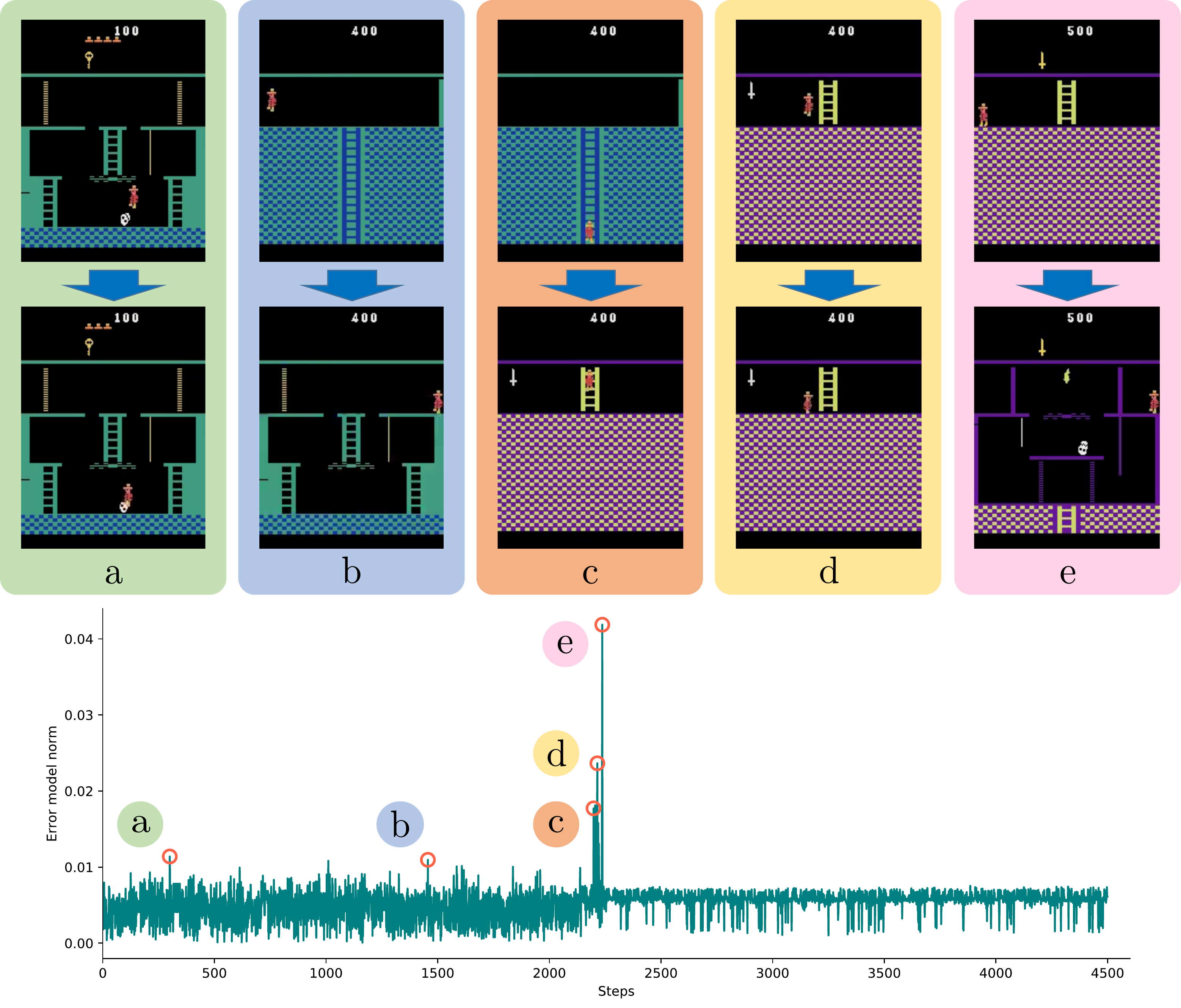}
    \caption{Evolution of the error model norm, and the five representative transitions with the high norm values, in a sample episode of the EMI agent on Montezuma's Revenge. The y-axis and the x-axis mean the error model norm and the step number in the episode, respectively. Each colored pair of two images represent $s_t$ (upper) and $s_t'$ (lower) of its corresponding transition. In transitions \textit{b}, \textit{c}, and \textit{e}, $s_t$ and $s_t'$ are from different rooms with distant background images. In transitions \textit{a} and \textit{d}, the agent is off the platform and thus has no control over itself in $s_t$.}
    \label{fig:reconciler_evolution}
    \vspace{-0.5em}
\end{figure*}
To get an understanding of the empirical behavior of the error model, we visualize the evolution of the error model norm $\|S(s_t, a_t)\|_2$ throughout a full episode from one of our experiments on Montezuma's Revenge, in \Cref{fig:reconciler_evolution}. We picked five representative transitions with high values of the error model norm from the episode. The upper and lower images of each transition in the figure represent $s_t$ and $s_t'$, respectively.

In the case of transitions \textit{b}, \textit{c}, and \textit{e}, due to the discrepancy between the two distinct background images, $\|\phi(s_t) - \phi(s_t')\|_2$ easily becomes large which makes the residual error as well as the error term larger, too. Transitions \textit{a} and \textit{d} belong to the case where the action chosen by the policy has no or almost no effect on $s_t'$ i.e. $P(s_t'|s_t, a_t) \approx P(s_t'|s_t)$. Linear models without any error terms can fail in such events easily. Thus, the error term in our model gets bigger to mitigate the modeling error.

In conclusion, we observed the error terms generally had much larger norms in the cases such as the representative transitions, in order to alleviate the occasional irreducible large residual errors under the linear dynamics model. 

\section{Convergence of loss terms}
\Cref{fig:loss_plots} shows the convergence of loss terms in Equation (7) from the main text. All loss terms reach convergence within the first 50 iterations, which verifies that EMI successfully learns desired embedding representations.

\section{Statistical tests}
As TRPO exhibits high-variance results, we ran more seeds to verify the statistical significance of EMI. We ran 15 random seeds on the SparseHalfCheetah environment which we claim EMI outperforms other baselines, the difference in the mean returns is relatively small, and the variance is high.
We then performed the t-test to confirm the statistical significance following the practice from \citet{colas2018gep}. For each baseline methods, we report t-values with p-values in parentheses. (Results are significant when p < 0.05)
\begin{itemize}
\item EMI vs ICM: 8.58 (2.99e-7)
\item EMI vs RND: 8.57 (2.96e-7)
\item EMI vs EX2: 1.81 (0.0410)
\end{itemize}

The results show that in SparseHalfCheetah environment, EMI outperforms the baseline methods within the 95\% confidence level.

\section{The BoxImage experiment}
The intrinsic position of the agent, $x$, is constrained within $x \in [0, 100]^2$. Observations the agent receives are $52 \times 52 \times 1$ images, each of which has a white circle that corresponds to the intrinsic position of the agent on a black background. The agent can move itself by performing an action $a \in [-1, 1]^2$. Concretely, if the agent performs $a$ at $x$, its next intrinsic position will be $\min(\max(x + a, (0, 0)), (100, 100))$. The initial intrinsic position of the agent, $x_i$, is randomly chosen satisfying $\|x_i\| \ge 75$.

We collected 30,000 samples with a randomly initialized TRPO policy in BoxImage. Using the above samples, we trained two set of embedding functions each with the same hyper-parameters ($\lambda_{\text{error}}=100, \lambda_{\text{info}}=0.01, d=2$) but with an exception of whether to regularize the \emph{state} or the \emph{action} embeddings.

\end{document}